\documentclass[10pt,twocolumn,letterpaper]{article}

\newif\ifreviewmode
\reviewmodefalse          %
\ifreviewmode
  \usepackage[review]{cvpr}   %
\else
  \usepackage{cvpr}                   %
\fi

\usepackage{multirow}
\usepackage{microtype}
\usepackage{array}           %
\usepackage{adjustbox}       %
\usepackage{xspace}          %
\usepackage{siunitx}         %
\usepackage{afterpage}       %
\usepackage{needspace}       %

\ifdefined\pdfsuppressptexinfo \pdfsuppressptexinfo=-1 \fi
\ifdefined\pdfinfoomitdate \pdfinfoomitdate=1 \fi
\ifdefined\pdftrailerid \pdftrailerid{} \fi

\graphicspath{{figs/}{generated/}}

\makeatletter
\ifx\ifreviewmode\@undefined
  \PackageError{preamble}{\expandafter\string\csname ifreviewmode\endcsname\space is not defined}%
    {Define the switch with \string\newif\expandafter\string\csname ifreviewmode\endcsname\space in the document before \string.}
\fi
\ifreviewmode
 \ifdefined\c@linenumber       %
  \ifdefined\pdfsavepos        %
    \@ifundefined{AddToShipoutPicture}{\usepackage{eso-pic}}{}
    \newcommand\LN@rec[4]{%
      \@ifundefined{LN@page@#4}{\expandafter\gdef\csname LN@page@#4\endcsname{}}{}%
      \expandafter\g@addto@macro\csname LN@page@#4\endcsname{\LN@draw{#1}{#2}{#3}}}
    \def\LN@mid{20054016}%
    \newcommand\LN@draw[3]{%
      \ifnum#2<\LN@mid\relax
        \put(\LenToUnit{#2sp},\LenToUnit{#3sp}){\makebox[0pt][r]{\linenumberfont#1\hspace{\linenumbersep}}}%
      \else
        \put(\LenToUnit{#2sp},\LenToUnit{#3sp}){\makebox[0pt][l]{\hspace{\columnwidth}\hspace{\linenumbersep}\linenumberfont#1}}%
      \fi}
    \def\LN@record{%
      \pdfsavepos
      \edef\@tempa{\write\@auxout{\string\LN@rec{\the\c@linenumber}{\noexpand\the\noexpand\pdflastxpos}%
        {\noexpand\the\noexpand\pdflastypos}{\noexpand\the\noexpand\c@page}}}\@tempa}
    \let\makeLineNumberRunning\LN@record
    \let\makeLineNumberOdd\LN@record
    \let\makeLineNumberEven\LN@record
    \AddToShipoutPicture{\AtPageLowerLeft{\csname LN@page@\the\c@page\endcsname}}
  \fi
 \fi
\fi
\makeatother

\makeatletter
\renewcommand\paragraph{\@startsection{paragraph}{4}{\z@}{1.5ex \@plus .5ex \@minus .2ex}{-1em}{\normalfont\normalsize\bfseries}}
\makeatother

\newcommand{\tbd}[1]{\textcolor{red}{\textbf{[TBD:} #1\textbf{]}}}

\newcommand{\phfig}[2]{\fbox{\parbox[c][#1][c]{\dimexpr\linewidth-2\fboxsep-2\fboxrule\relax}{\centering\tbd{FIGURE #2}}}}

\providecommand{\dryvalue}[1]{\textcolor{red}{#1}}
\definecolor{confirmedgreen}{RGB}{0,128,0}
\providecommand{\okvalue}[1]{\textcolor{confirmedgreen}{#1}}

\newif\ifsrcnotes \srcnotesfalse

\IfFileExists{generated/config_macros.tex}{  %
}{}   %
\IfFileExists{generated/macros.tex}{\providecommand{\dryvalue}[1]{\textcolor{red}{#1}}

\newcommand{\NScenes}{\okvalue{24}}  %
\newcommand{\VlmRejectedPrec}{\okvalue{80}}  %
\newcommand{\VlmFixedShare}{\dryvalue{81.9}}  %
\newcommand{\VlmBrokeShare}{\dryvalue{16.8}}  %
\newcommand{\VlmFixedTrain}{\dryvalue{16.0K}}  %
\newcommand{\VlmBrokeTrain}{\dryvalue{3.3K}}  %
\newcommand{\VlmRejectedTrain}{\dryvalue{161K}}  %
\newcommand{\TwoDTypeAccPct}{\dryvalue{96.2}}  %
\newcommand{\TwoDmAPNoVlm}{\okvalue{0.304}}  %
\newcommand{\TwoDmAPVlm}{\okvalue{0.337}}  %
\newcommand{\ReviewFlagCatch}{\okvalue{25}}  %
\newcommand{\DhakaBKeyframes}{\okvalue{1,196}}  %
\newcommand{\DhakaBRowsPerImage}{\okvalue{9.20}}  %
\newcommand{\PAArmBLocal}{\okvalue{0.860}}  %
\newcommand{\PAMergeLocal}{\okvalue{0.791}}  %
\newcommand{\LoraNusVanAcc}{\okvalue{86.9}}  %
\newcommand{\LoraNusFtAcc}{\okvalue{95.0}}  %
\newcommand{\LoraVerVanAcc}{\okvalue{52.5}}  %
\newcommand{\LoraVerFtAcc}{\okvalue{82.5}}  %
\newcommand{\DVVespaBoxes}{\okvalue{13.5K}}  %
\newcommand{\DVHestiaBoxes}{\okvalue{14.1K}}  %
\newcommand{\DVVespaLegacyAuto}{\okvalue{4.0}}  %
\newcommand{\DVHestiaLegacyAuto}{\okvalue{40.4}}  %
\newcommand{\DhakaKeyframes}{\okvalue{400}}  %
\newcommand{\DhakaRowsPerImage}{\okvalue{11.98}}  %
\newcommand{\NusRowsPerImage}{\okvalue{3.51}}  %
\newcommand{\DhakaArmBShare}{\okvalue{20.4}}  %
}{}                 %

\IfFileExists{generated/results_macros.tex}{  %
\newcommand{\FullValScenes}{\okvalue{150}}  %
\newcommand{\FullValKeyframes}{\okvalue{6,019}}  %
\newcommand{\DetGTOnemAP}{\okvalue{0.820}}  %
\newcommand{\DetGTOneNDS}{\okvalue{0.722}}  %
\newcommand{\DetGTThreemAP}{\okvalue{0.769}}  %
\newcommand{\DetGTThreeNDS}{\okvalue{0.693}}  %
\newcommand{\DetGTEightmAP}{\okvalue{0.571}}  %
\newcommand{\DetGTEightNDS}{\okvalue{0.572}}  %
\newcommand{\TrainLabelScenes}{\okvalue{700}}  %
\newcommand{\TrainLabelHestiaBoxes}{\okvalue{1.48M}}  %
\newcommand{\QuotedVespaCPOnemAP}{\okvalue{0.530}}  %
\newcommand{\QuotedVespaCPOneNDS}{\okvalue{0.481}}  %
\newcommand{\QuotedVespaPLThreemAP}{\okvalue{0.222}}  %
\newcommand{\QuotedVespaCPThreemAP}{\okvalue{0.465}}  %
\newcommand{\QuotedVespaPLThreeNDS}{\okvalue{0.230}}  %
\newcommand{\QuotedVespaCPThreeNDS}{\okvalue{0.435}}  %
\newcommand{\QuotedVespaPLEightmAP}{\okvalue{0.110}}  %
\newcommand{\QuotedVespaCPEightmAP}{\okvalue{0.248}}  %
\newcommand{\QuotedVespaPLEightNDS}{\okvalue{0.147}}  %
\newcommand{\QuotedVespaCPEightNDS}{\okvalue{0.290}}  %
\newcommand{\QuotedUnionCPOnemAP}{\okvalue{0.384}}  %
\newcommand{\QuotedUnionCPOneNDS}{\okvalue{0.312}}  %
\newcommand{\QuotedUnionPLThreemAP}{\okvalue{0.073}}  %
\newcommand{\QuotedUnionCPThreemAP}{\okvalue{0.272}}  %
\newcommand{\QuotedUnionPLThreeNDS}{\okvalue{0.110}}  %
\newcommand{\QuotedUnionCPThreeNDS}{\okvalue{0.252}}  %
\newcommand{\QuotedUnionPLEightmAP}{\okvalue{0.033}}  %
\newcommand{\QuotedUnionCPEightmAP}{\okvalue{0.120}}  %
\newcommand{\QuotedUnionPLEightNDS}{\okvalue{0.061}}  %
\newcommand{\QuotedUnionCPEightNDS}{\okvalue{0.130}}  %
\newcommand{\QuotedAnnoFreeCPOnemAP}{\okvalue{0.455}}  %
\newcommand{\QuotedAnnoFreeCPOneNDS}{\okvalue{0.358}}  %
\newcommand{\QuotedAnnoFreeCPThreemAP}{\okvalue{0.344}}  %
\newcommand{\QuotedAnnoFreeCPThreeNDS}{\okvalue{0.366}}  %
\newcommand{\QuotedAnnoFreeCPEightmAP}{\okvalue{0.244}}  %
\newcommand{\QuotedAnnoFreeCPEightNDS}{\okvalue{0.264}}  %
\newcommand{\QuotedCMCPOnemAP}{\okvalue{0.279}}  %
\newcommand{\QuotedCMCPOneNDS}{\okvalue{0.276}}  %
\newcommand{\DkDetHestiaOnemAP}{\okvalue{0.342}}  %
\newcommand{\DkDetHestiaOneNDS}{\okvalue{0.349}}  %
\newcommand{\DkDetHestiaThreemAP}{\okvalue{0.278}}  %
\newcommand{\DkDetHestiaThreeNDS}{\okvalue{0.344}}  %
\newcommand{\DkDetHestiaEightmAP}{\okvalue{0.124}}  %
\newcommand{\DkDetHestiaEightNDS}{\okvalue{0.257}}  %
\newcommand{\DkDetHestiaAPCycleRickshaw}{\okvalue{0.137}}  %
\newcommand{\DkDetHestiaAPAutoRickshaw}{\okvalue{0.094}}  %
\newcommand{\DkDetVespaOnemAP}{\okvalue{0.215}}  %
\newcommand{\DkDetVespaOneNDS}{\okvalue{0.177}}  %
\newcommand{\DkDetVespaThreemAP}{\okvalue{0.138}}  %
\newcommand{\DkDetVespaThreeNDS}{\okvalue{0.145}}  %
\newcommand{\DkDetVespaEightmAP}{\okvalue{0.053}}  %
\newcommand{\DkDetVespaEightNDS}{\okvalue{0.105}}  %
\newcommand{\DkDetVespaAPCycleRickshaw}{\dryvalue{0.037}}  %
\newcommand{\DkDetVespaAPAutoRickshaw}{\dryvalue{0.008}}  %
\newcommand{\DkTrainScenes}{\okvalue{140}}  %
\newcommand{\DkTestScenes}{\okvalue{30}}  %
\newcommand{\DkTrainKeyframes}{\okvalue{13.8K}}  %
\newcommand{\DkTestKeyframes}{\okvalue{2,963}}  %
\newcommand{\DkHestiaTrainBoxes}{\okvalue{485K}}  %
\newcommand{\DkVespaTrainBoxes}{\okvalue{351K}}  %
\newcommand{\DetHestiaVtwoOnemAP}{\okvalue{0.664}}  %
\newcommand{\DetHestiaVtwoOneNDS}{\okvalue{0.577}}  %
\newcommand{\DetHestiaVtwoThreemAP}{\okvalue{0.599}}  %
\newcommand{\TrainLabelHestiaVtwoThreemAP}{\okvalue{0.529}}  %
\newcommand{\DetHestiaVtwoThreeNDS}{\okvalue{0.539}}  %
\newcommand{\TrainLabelHestiaVtwoThreeNDS}{\okvalue{0.574}}  %
\newcommand{\DetHestiaVtwoEightmAP}{\okvalue{0.360}}  %
\newcommand{\TrainLabelHestiaVtwoEightmAP}{\okvalue{0.337}}  %
\newcommand{\DetHestiaVtwoEightNDS}{\okvalue{0.378}}  %
\newcommand{\TrainLabelHestiaVtwoEightNDS}{\okvalue{0.417}}  %
\newcommand{\FullHestiaVtwoDevmAOE}{\okvalue{0.741}}  %
\newcommand{\FullHestiaVtwoDevmAOEPreYaw}{\okvalue{0.928}}  %
\newcommand{\FullHestiaVtwoDevNDSPreYaw}{\okvalue{0.401}}  %
\newcommand{\FullHestiaVtwoDevNDS}{\okvalue{0.420}}  %
\newcommand{\WmNusRefN}{\okvalue{134K}}  %
\newcommand{\WmNusRefInputIou}{\okvalue{0.293}}  %
\newcommand{\WmNusRefInputAcc}{\okvalue{21.8}}  %
\newcommand{\WmNusRefInputCtr}{\okvalue{0.54}}  %
\newcommand{\WmNusRefInputYaw}{\okvalue{15.2}}  %
\newcommand{\WmNusRefWIou}{\okvalue{0.317}}  %
\newcommand{\WmNusRefWAcc}{\okvalue{27.5}}  %
\newcommand{\WmNusRefWCtr}{\okvalue{0.50}}  %
\newcommand{\WmNusRefWYaw}{\okvalue{16.2}}  %
\newcommand{\WmNusYawWFinalMed}{\okvalue{3.43}}  %
\newcommand{\WmNusYawWFinalFlip}{\okvalue{6.2}}  %
\newcommand{\TvFusAMOTA}{\dryvalue{0.365}}  %
\newcommand{\TvFusMOTA}{\dryvalue{0.312}}  %
\newcommand{\TvFusRecall}{\dryvalue{0.518}}  %
\newcommand{\TvFusIDS}{\dryvalue{3,410}}  %
\newcommand{\TvFusFRAG}{\dryvalue{1,086}}  %
\newcommand{\TvFusMT}{\dryvalue{2,263}}  %
\newcommand{\TvFusAMOTABus}{\dryvalue{0.393}}  %
\newcommand{\TvFusAMOTATruck}{\dryvalue{0.148}}  %
\newcommand{\TvNoFusAMOTA}{\dryvalue{0.340}}  %
\newcommand{\TvNoFusMOTA}{\dryvalue{0.290}}  %
\newcommand{\TvNoFusRecall}{\dryvalue{0.479}}  %
\newcommand{\TvNoFusIDS}{\dryvalue{3,136}}  %
\newcommand{\TvNoFusFRAG}{\dryvalue{1,441}}  %
\newcommand{\TvNoFusMT}{\dryvalue{2,070}}  %
\newcommand{\TvNoFusAMOTABus}{\dryvalue{0.324}}  %
\newcommand{\TvNoFusAMOTATruck}{\dryvalue{0.114}}  %
\newcommand{\TvSamAMOTA}{\dryvalue{0.257}}  %
\newcommand{\TvSamMOTA}{\dryvalue{0.232}}  %
\newcommand{\TvSamRecall}{\dryvalue{0.422}}  %
\newcommand{\TvSamIDS}{\dryvalue{6,222}}  %
\newcommand{\TvSamFRAG}{\dryvalue{1,598}}  %
\newcommand{\TvSamMT}{\dryvalue{1,524}}  %
\newcommand{\TtFusAMOTA}{\dryvalue{0.316}}  %
\newcommand{\TtFusMOTA}{\dryvalue{0.269}}  %
\newcommand{\TtFusRecall}{\dryvalue{0.444}}  %
\newcommand{\TtFusIDS}{\dryvalue{15.7K}}  %
\newcommand{\TtFusFRAG}{\dryvalue{5,589}}  %
\newcommand{\TtFusMT}{\dryvalue{9,867}}  %
\newcommand{\TtNoFusAMOTA}{\dryvalue{0.291}}  %
\newcommand{\TtNoFusMOTA}{\dryvalue{0.249}}  %
\newcommand{\TtNoFusRecall}{\dryvalue{0.424}}  %
\newcommand{\TtNoFusIDS}{\dryvalue{15.7K}}  %
\newcommand{\TtNoFusFRAG}{\dryvalue{7,731}}  %
\newcommand{\TtNoFusMT}{\dryvalue{9,230}}  %
\newcommand{\TtSamAMOTA}{\dryvalue{0.206}}  %
\newcommand{\TtSamMOTA}{\dryvalue{0.189}}  %
\newcommand{\TtSamRecall}{\dryvalue{0.381}}  %
\newcommand{\TtSamIDS}{\dryvalue{35.6K}}  %
\newcommand{\TtSamFRAG}{\dryvalue{9,355}}  %
\newcommand{\TtSamMT}{\dryvalue{7,117}}  %
\newcommand{\DetHestiaVtwoOneSharemAP}{\okvalue{80.9}}  %
\newcommand{\TvFusIDSRate}{\dryvalue{6.1}}  %
\newcommand{\TvNoFusIDSRate}{\dryvalue{6.1}}  %
\newcommand{\TtFusIDSRate}{\dryvalue{5.9}}  %
\newcommand{\TtNoFusIDSRate}{\dryvalue{6.1}}  %
\newcommand{\BestQuotedUnionPLThreemAP}{\QuotedUnionPLThreemAP}
\newcommand{\BestQuotedVespaPLThreemAP}{\QuotedVespaPLThreemAP}
\newcommand{\BestTrainLabelHestiaVtwoThreemAP}{\textbf{\TrainLabelHestiaVtwoThreemAP}}
\newcommand{\BestQuotedUnionPLThreeNDS}{\QuotedUnionPLThreeNDS}
\newcommand{\BestQuotedVespaPLThreeNDS}{\QuotedVespaPLThreeNDS}
\newcommand{\BestTrainLabelHestiaVtwoThreeNDS}{\textbf{\TrainLabelHestiaVtwoThreeNDS}}
\newcommand{\BestQuotedUnionPLEightmAP}{\QuotedUnionPLEightmAP}
\newcommand{\BestQuotedVespaPLEightmAP}{\QuotedVespaPLEightmAP}
\newcommand{\BestTrainLabelHestiaVtwoEightmAP}{\textbf{\TrainLabelHestiaVtwoEightmAP}}
\newcommand{\BestQuotedUnionPLEightNDS}{\QuotedUnionPLEightNDS}
\newcommand{\BestQuotedVespaPLEightNDS}{\QuotedVespaPLEightNDS}
\newcommand{\BestTrainLabelHestiaVtwoEightNDS}{\textbf{\TrainLabelHestiaVtwoEightNDS}}
\newcommand{\BestQuotedUnionCPOnemAP}{\QuotedUnionCPOnemAP}
\newcommand{\BestQuotedCMCPOnemAP}{\QuotedCMCPOnemAP}
\newcommand{\BestQuotedAnnoFreeCPOnemAP}{\QuotedAnnoFreeCPOnemAP}
\newcommand{\BestQuotedVespaCPOnemAP}{\QuotedVespaCPOnemAP}
\newcommand{\BestDetHestiaVtwoOnemAP}{\textbf{\DetHestiaVtwoOnemAP}}
\newcommand{\BestQuotedUnionCPOneNDS}{\QuotedUnionCPOneNDS}
\newcommand{\BestQuotedCMCPOneNDS}{\QuotedCMCPOneNDS}
\newcommand{\BestQuotedAnnoFreeCPOneNDS}{\QuotedAnnoFreeCPOneNDS}
\newcommand{\BestQuotedVespaCPOneNDS}{\QuotedVespaCPOneNDS}
\newcommand{\BestDetHestiaVtwoOneNDS}{\textbf{\DetHestiaVtwoOneNDS}}
\newcommand{\BestQuotedUnionCPThreemAP}{\QuotedUnionCPThreemAP}

\newcommand{\BestQuotedAnnoFreeCPThreemAP}{\QuotedAnnoFreeCPThreemAP}
\newcommand{\BestQuotedVespaCPThreemAP}{\QuotedVespaCPThreemAP}
\newcommand{\BestDetHestiaVtwoThreemAP}{\textbf{\DetHestiaVtwoThreemAP}}
\newcommand{\BestQuotedUnionCPThreeNDS}{\QuotedUnionCPThreeNDS}

\newcommand{\BestQuotedAnnoFreeCPThreeNDS}{\QuotedAnnoFreeCPThreeNDS}
\newcommand{\BestQuotedVespaCPThreeNDS}{\QuotedVespaCPThreeNDS}
\newcommand{\BestDetHestiaVtwoThreeNDS}{\textbf{\DetHestiaVtwoThreeNDS}}
\newcommand{\BestQuotedUnionCPEightmAP}{\QuotedUnionCPEightmAP}

\newcommand{\BestQuotedAnnoFreeCPEightmAP}{\QuotedAnnoFreeCPEightmAP}
\newcommand{\BestQuotedVespaCPEightmAP}{\QuotedVespaCPEightmAP}
\newcommand{\BestDetHestiaVtwoEightmAP}{\textbf{\DetHestiaVtwoEightmAP}}
\newcommand{\BestQuotedUnionCPEightNDS}{\QuotedUnionCPEightNDS}

\newcommand{\BestQuotedAnnoFreeCPEightNDS}{\QuotedAnnoFreeCPEightNDS}
\newcommand{\BestQuotedVespaCPEightNDS}{\QuotedVespaCPEightNDS}
\newcommand{\BestDetHestiaVtwoEightNDS}{\textbf{\DetHestiaVtwoEightNDS}}
\newcommand{\BestWmNusRefInputIou}{\WmNusRefInputIou}
\newcommand{\BestWmNusRefWIou}{\textbf{\WmNusRefWIou}}
\newcommand{\BestWmNusRefInputAcc}{\WmNusRefInputAcc}
\newcommand{\BestWmNusRefWAcc}{\textbf{\WmNusRefWAcc}}
\newcommand{\BestWmNusRefInputCtr}{\WmNusRefInputCtr}
\newcommand{\BestWmNusRefWCtr}{\textbf{\WmNusRefWCtr}}
\newcommand{\BestWmNusRefInputYaw}{\textbf{\WmNusRefInputYaw}}
\newcommand{\BestWmNusRefWYaw}{\WmNusRefWYaw}
\newcommand{\BestFullHestiaVtwoDevNDSPreYaw}{\FullHestiaVtwoDevNDSPreYaw}
\newcommand{\BestFullHestiaVtwoDevNDS}{\textbf{\FullHestiaVtwoDevNDS}}
\newcommand{\BestFullHestiaVtwoDevmAOEPreYaw}{\FullHestiaVtwoDevmAOEPreYaw}
\newcommand{\BestFullHestiaVtwoDevmAOE}{\textbf{\FullHestiaVtwoDevmAOE}}
\newcommand{\BestDkDetVespaOnemAP}{\DkDetVespaOnemAP}
\newcommand{\BestDkDetHestiaOnemAP}{\textbf{\DkDetHestiaOnemAP}}
\newcommand{\BestDkDetVespaOneNDS}{\DkDetVespaOneNDS}
\newcommand{\BestDkDetHestiaOneNDS}{\textbf{\DkDetHestiaOneNDS}}
\newcommand{\BestDkDetVespaThreemAP}{\DkDetVespaThreemAP}
\newcommand{\BestDkDetHestiaThreemAP}{\textbf{\DkDetHestiaThreemAP}}
\newcommand{\BestDkDetVespaThreeNDS}{\DkDetVespaThreeNDS}
\newcommand{\BestDkDetHestiaThreeNDS}{\textbf{\DkDetHestiaThreeNDS}}
\newcommand{\BestDkDetVespaEightmAP}{\DkDetVespaEightmAP}
\newcommand{\BestDkDetHestiaEightmAP}{\textbf{\DkDetHestiaEightmAP}}
\newcommand{\BestDkDetVespaEightNDS}{\DkDetVespaEightNDS}
\newcommand{\BestDkDetHestiaEightNDS}{\textbf{\DkDetHestiaEightNDS}}
\newcommand{\BestTvFusAMOTA}{\textbf{\TvFusAMOTA}}
\newcommand{\BestTvNoFusAMOTA}{\TvNoFusAMOTA}
\newcommand{\BestTvSamAMOTA}{\TvSamAMOTA}
\newcommand{\BestTtFusAMOTA}{\textbf{\TtFusAMOTA}}
\newcommand{\BestTtNoFusAMOTA}{\TtNoFusAMOTA}
\newcommand{\BestTtSamAMOTA}{\TtSamAMOTA}
\newcommand{\BestTvFusMOTA}{\textbf{\TvFusMOTA}}
\newcommand{\BestTvNoFusMOTA}{\TvNoFusMOTA}
\newcommand{\BestTvSamMOTA}{\TvSamMOTA}
\newcommand{\BestTtFusMOTA}{\textbf{\TtFusMOTA}}
\newcommand{\BestTtNoFusMOTA}{\TtNoFusMOTA}
\newcommand{\BestTtSamMOTA}{\TtSamMOTA}
\newcommand{\BestTvFusRecall}{\textbf{\TvFusRecall}}
\newcommand{\BestTvNoFusRecall}{\TvNoFusRecall}
\newcommand{\BestTvSamRecall}{\TvSamRecall}
\newcommand{\BestTtFusRecall}{\textbf{\TtFusRecall}}
\newcommand{\BestTtNoFusRecall}{\TtNoFusRecall}
\newcommand{\BestTtSamRecall}{\TtSamRecall}
\newcommand{\BestTvFusIDS}{\TvFusIDS}
\newcommand{\BestTvNoFusIDS}{\textbf{\TvNoFusIDS}}
\newcommand{\BestTvSamIDS}{\TvSamIDS}
\newcommand{\BestTtFusIDS}{\TtFusIDS}
\newcommand{\BestTtNoFusIDS}{\TtNoFusIDS}
\newcommand{\BestTtSamIDS}{\TtSamIDS}
\newcommand{\BestTvFusFRAG}{\textbf{\TvFusFRAG}}
\newcommand{\BestTvNoFusFRAG}{\TvNoFusFRAG}
\newcommand{\BestTvSamFRAG}{\TvSamFRAG}
\newcommand{\BestTtFusFRAG}{\textbf{\TtFusFRAG}}
\newcommand{\BestTtNoFusFRAG}{\TtNoFusFRAG}
\newcommand{\BestTtSamFRAG}{\TtSamFRAG}
\newcommand{\BestTvFusMT}{\textbf{\TvFusMT}}
\newcommand{\BestTvNoFusMT}{\TvNoFusMT}
\newcommand{\BestTvSamMT}{\TvSamMT}
\newcommand{\BestTtFusMT}{\textbf{\TtFusMT}}
\newcommand{\BestTtNoFusMT}{\TtNoFusMT}
\newcommand{\BestTtSamMT}{\TtSamMT}
}{}  %
\IfFileExists{macro_fallbacks.tex}{\providecommand{\NScenes}{\tbd{NScenes}}

\providecommand{\VlmRejectedPrec}{\tbd{VlmRejectedPrec}}

\providecommand{\TwoDmAPNoVlm}{\tbd{TwoDmAPNoVlm}}
\providecommand{\TwoDmAPVlm}{\tbd{TwoDmAPVlm}}

\providecommand{\ReviewFlagCatch}{\tbd{ReviewFlagCatch}}

\providecommand{\DhakaKeyframes}{\tbd{DhakaKeyframes}}

\providecommand{\DhakaRowsPerImage}{\tbd{DhakaRowsPerImage}}
\providecommand{\NusRowsPerImage}{\tbd{NusRowsPerImage}}

\providecommand{\DhakaArmBShare}{\tbd{DhakaArmBShare}}

}{}                  %

\newcommand{\sref}[1]{\ifcsname sref@#1\endcsname\csname sref@#1\endcsname\else\tbd{supplement reference #1}\fi}

\newcommand{\hestia}{HESTIA\xspace}           %
\newcommand{\vespa}{VESPA\xspace}             %
\newcommand{\dhakadata}{DHAKA200\xspace}     %
\newcommand{\DkAllScenes}{200}
\newcommand{\DkValScenes}{30}
\newcommand{\DkSceneSec}{20}
\newcommand{\DkKeyframeHz}{5}

\newcommand{\armA}{arm A\xspace}              %
\newcommand{\armB}{arm B\xspace}              %
\newcommand{\ArmA}{Arm A\xspace}              %
\newcommand{\ArmB}{Arm B\xspace}
\newcommand{\gate}{bearing gate\xspace}
\newcommand{\verifier}{VLM verifier\xspace}
\newcommand{\refiner}{box refiner\xspace}
\newcommand{\yawmodel}{yaw model\xspace}
\newcommand{\fusion}{temporal fusion\xspace}
\newcommand{\map}{mAP\xspace}
\newcommand{\nds}{NDS\xspace}
\newcommand{\iou}{3D IoU\xspace}
\newcommand{\bearing}{\ensuremath{\theta}}    %
\newcommand{\yaw}{\ensuremath{\psi}}          %
\newcommand{\etype}[1]{\textit{#1}}           %

\definecolor{linkblue}{rgb}{0.21,0.49,0.74}
\usepackage[pagebackref,breaklinks,colorlinks,allcolors=linkblue]{hyperref}
\hypersetup{pdfauthor={},pdftitle={}}    %

\def\confName{CVPR}
\def\confYear{2027}

\title{HESTIA: Temporally Consistent 3D Annotation of Dense Urban Traffic with Vision Language Verification}

\ifreviewmode
  \author{Anonymous CVPR 2027 submission}   %
\else
  \author{Zamiul Rashid$^{1}$ \quad Mollah Md Saif$^{1}$ \quad Hironmoy Roy Rudra$^{1}$ \quad Debjoty Bhowmik$^{1}$\\
Md.\ Rafid Khan$^{1}$ \quad Shahriar Hossain$^{2}$ \quad Md Khalilur Rhaman$^{1}$ \quad Md.\ Golam Rabiul Alam$^{1}$\\[2pt]
$^{1}$BRAC University \qquad $^{2}$George Mason University\\[2pt]
{\small\tt zamiul.rashid@g.bracu.ac.bd, mollah.mdsaif@bracu.ac.bd, hironmoy.roy.rudra@g.bracu.ac.bd,}\\
{\small\tt debjoty.bhowmik@g.bracu.ac.bd, md.rafid.khan@g.bracu.ac.bd, shossa4@gmu.edu,}\\
{\small\tt khalilur@bracu.ac.bd, rabiul.alam@bracu.ac.bd}}
\fi

\begin{document}
\maketitle
\renewcommand{\okvalue}[1]{#1}
\renewcommand{\dryvalue}[1]{#1}
\begin{abstract}
Hand labeling 3D boxes for dense traffic is slow, and South~Asian traffic adds vehicle classes unknown to public detectors.
We present \hestia, an annotation pipeline that decides in camera images which objects exist and lets LiDAR place them.
Two merged 2D detectors, a bearing gate, Segment Anything Model (SAM) tracking and a blind vision language model (VLM) verifier produce masks and tracks.
A lift, a box refiner, temporal fusion and a yaw model turn them into 3D boxes.
The refiner and yaw model are trained on Waymo Open labels.
On the nuScenes train split the \hestia pseudo labels reach \TrainLabelHestiaVtwoEightmAP{} eight class mean average precision (mAP).
\vespa trains on no label and reports \QuotedVespaPLEightmAP{} there, and every box of both labelers carries a confidence score of one.
A CenterPoint 3D detector trained on these labels scores \DetHestiaVtwoOnemAP{} class agnostic mAP, against \QuotedVespaCPOnemAP{} with \vespa labels.
On nuScenes val the smoothing of temporal fusion raises tracking accuracy AMOTA from \TvNoFusAMOTA{} to \TvFusAMOTA{}.
The second 2D detector is a YOLO11x fine tuned on South Asian road datasets and adds cycle rickshaws and auto rickshaws.
The VLM verifier is fine tuned on driving and South Asian datasets.
We release \dhakadata with our 3D boxes and instance masks.
It holds two hundred Dhaka scenes in nuScenes format.
Every city that a self-driving or driver-assist system enters brings its own traffic mix and vehicle types.
An automatic 3D labeler lets a team build training data for that city without months of hand labeling.
\end{abstract}
\section{Introduction}
\begin{figure}[t]
\centering
\IfFileExists{figs/fig_teaser.pdf}{\includegraphics[width=\linewidth]{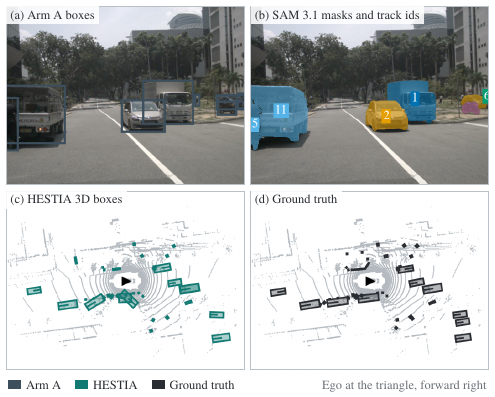}}{\phfig{3.1in}{fig\_teaser}}
\caption{\textbf{From image masks to 3D pseudo labels:} \hestia decides in the camera images which objects exist and what they are.
It then lifts each mask into LiDAR, where the \refiner fits the box and \fusion smooths it over time.
Panel (a) shows the boxes of the common class detector \armA.
Panel (b) shows the SAM~3.1 masks with their track ids.
Panels (c) and (d) show the \hestia boxes and the ground truth boxes from above.
The boxes come from the val subset run with an earlier \refiner and \yawmodel.}
\label{fig:teaser}
\end{figure}
\label{sec:intro}

Training a 3D detector needs boxes with a class, a heading and a track identity per object and keyframe.
Labeling them by hand is slow, and dense traffic slows it further because objects overlap in the image.
Offboard labelers cut this cost, but they still train on human labels from the target domain~\cite{qi2021offboard,ma2023detzero}.
A newer family of labelers removes that need and uses no human label at all.
These labelers lift the output of 2D foundation models into LiDAR and fit boxes there~\cite{tempfli2026vespa,lee2025openbox,ma2026autoexpert}.

\vespa is a recent labeler of that family with public code~\cite{tempfli2026vespa}.
It scores above the other quoted labelers of \cref{tab:protocol}.
An open vocabulary detector finds objects in the images from class names.
The Segment Anything Model (SAM) segments them, and the masks are lifted into the LiDAR sweeps.
Clusters of points become boxes under size priors that a language model supplies, and every box gets the same confidence score of one.
A CenterPoint detector trained on \vespa pseudo labels reaches about two thirds of the supervised class agnostic \map.
\vespa needs no training and no human label, which makes it the natural baseline.
\vespa names only its prompted classes and fits every box under a fixed size prior, with no learned correction.
To our knowledge no published labeler has been tested on dense South Asian traffic.

\begin{figure*}[t]
\centering
\IfFileExists{figs/fig_pipeline.pdf}{\includegraphics[width=\textwidth]{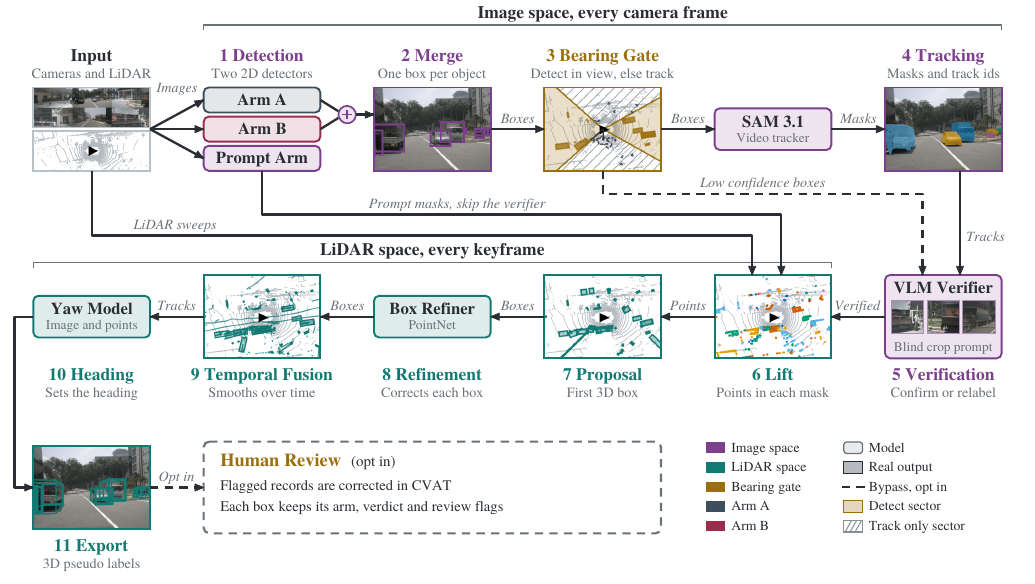}}{\phfig{3.3in}{fig\_pipeline}}
\caption{\textbf{Pipeline:} Two mask sources feed the lift, the verified tracks of the image stages and the SAM~3.1 prompt arm.
In the image stages two detectors are merged by class authority, and the bearing gate routes each box to detection or tracking. \ArmB is off in the nuScenes runs, where \armA detects alone.
SAM~3.1 tracks objects in each camera, stitching joins them across cameras and the VLM verifier judges each track or low confidence box.
Prompt arm masks and boxes under the crop floor skip the verifier.
Masks are lifted into the LiDAR, fitted, refined and fused over time, and the yaw model sets the heading.
A human reviews the per object record in CVAT as an opt in step.
The pictures are real outputs of one val keyframe from the val subset run.}
\label{fig:pipeline}
\end{figure*}

\hestia (Hybrid Expert Semi automatic Traffic Instance Annotation) is our camera first labeler for that traffic (\cref{fig:teaser}).
Camera first means that the images decide which objects exist and what they are, and LiDAR then fixes where they are.
A lift, a learned \refiner, \fusion and a \yawmodel turn image masks into 3D pseudo labels (\cref{fig:pipeline}).
The masks come from the \hestia image stages or from the \emph{prompt arm}, which prompts SAM~3.1 with class names~\cite{carion2025sam3,meta2026sam31}.
Unlike \vespa, \hestia has two trained 3D models, the \refiner and the \yawmodel.
We train both models on the Waymo Open dataset~\cite{sun2019scalability}, and no nuScenes label enters their training.

We compare with \vespa on nuScenes and on our dense Dhaka traffic sample \dhakadata.
No step reads the ground truth (GT) of a scene while labeling it.
The size priors, the export settings and the detector score floor were set with GT (\cref{sec:setup}).
Both labelers are scored in mean average precision (\map) with the evaluator of \vespa.
Every box of both labelers carries a confidence score of one.
The nuScenes runs use six cameras and no local class detector, as nuScenes has no rickshaws.
In this research, our contributions are as follows:
\needspace{3\baselineskip}
\begin{itemize}
  \item \textbf{The HESTIA Annotation Pipeline:} Two 2D detectors merged by class authority, a \gate, stitched SAM~3.1 tracking and a blind vision language model (VLM) verifier decide which objects exist. The second detector covers the local classes and is off on nuScenes.
  A lift, a point network \refiner, \fusion and a \yawmodel then turn the masks into 3D boxes.
  On the \TrainLabelScenes{} nuScenes train scenes the \hestia pseudo labels reach \TrainLabelHestiaVtwoEightmAP{} eight class \map against the GT (\cref{tab:protocol}).
  \vespa reports \QuotedVespaPLEightmAP{} for its pseudo labels there.
  A CenterPoint LiDAR detector~\cite{yin2021center} trained on the \hestia pseudo labels scores \DetHestiaVtwoOnemAP{} class agnostic \map on val, against \QuotedVespaCPOnemAP{} on \vespa pseudo labels.
  \item \textbf{Local Classes of Dense South Asian Traffic:} A second 2D detector trained on the RSUD20K road dataset and the Indian Driving Dataset (IDD) adds cycle rickshaws and auto rickshaws (\cref{sec:method_finetune}).
  The \verifier is fine tuned on crops of driving and South Asian datasets, and SAM~3.1 runs as released.
  On \dhakadata we train CenterPoint on \hestia labels and on \vespa labels and compare the two detectors (\cref{tab:dhakasense}).
  \item \textbf{Labels for Dense Dhaka Traffic:} We release \dhakadata with its \hestia labels under CC~BY-NC~4.0.
  Vehicles and persons carry 3D boxes with class, heading and size, plus instance masks in all six cameras.
  Each box records its detector arm, its \verifier verdict and its review flags (\cref{sec:results}), and no annotator has verified these labels.
\end{itemize}

\section{Related Work}
\label{sec:related}

\paragraph{Offboard and Automatic 3D Labeling:}
Offboard detection runs over whole LiDAR sequences and refines each object track with learned models~\cite{qi2021offboard}.
DetZero, CTRL and LabelFormer refine boxes or whole trajectories over long sequences~\cite{ma2023detzero,fan2023once,yang2023labelformer}.
\vespa is camera first and proposes objects with image masks and lifts them into open vocabulary LiDAR pseudo labels under VLM guidance~\cite{tempfli2026vespa}.
UNION clusters LiDAR by motion and appearance and CM3D lifts open vocabulary masks, both without human labels~\cite{lentsch2024union,khurana2024cm3d}.
OpenBox lifts 2D foundation model instances into LiDAR and fits boxes by rigidity and motion~\cite{lee2025openbox}.
AutoExpert detects and segments in RGB from expert guidelines and fits cuboids under size priors from a VLM~\cite{ma2026autoexpert}.
Segment, Lift and Fit and UniLiPs lift 2D cues into LiDAR and fit shapes or temporal pseudo labels~\cite{li2024segment,ghilotti2026unilips}.
Concurrent work prompts SAM~3 on the six nuScenes cameras and fits LiDAR boxes without training~\cite{deniz2026open}.
It reports devkit results on the full val split, and \cref{tab:learned} gives devkit scores of our labels there without a matched comparison.
\hestia follows the camera first recipe of this group, but its \refiner and \yawmodel are trained on 3D labels, here those of Waymo Open.

\paragraph{Promptable Segmentation and Tracking:}
Segment Anything and SAM~2 introduced promptable image and video segmentation~\cite{kirillov2023segment,ravi2024sam2}, and SAM~3 adds concept prompts that detect and track in video~\cite{carion2025sam3}.
Grounded SAM and SAM2MOT start masks and tracks from detector boxes~\cite{ren2024grounded,jiang2026sam2mot}.

\paragraph{Vision Language Verification:}
AutoVDC asks VLMs to judge flagged annotation errors and reports weak zero shot recall on localization noise~\cite{vasa2026autovdc}.
ClipGrader trains a CLIP based grader that scores the correctness of 2D boxes~\cite{lu2025clipgrader}.
VLM-PL and AIDE place a VLM in the loop of pseudo labeling and data engines~\cite{kim2024vlm,liang2024aide}.
Another pipeline validates pseudo LiDAR 3D annotation with a VLM~\cite{ma2026efficient}, and a VLM can size 3D boxes from image crops~\cite{chen2026improving}.
These works judge 2D labels or size 3D boxes, while the \verifier judges the class of each track before the 3D stages lift it.

\paragraph{Lifting and Box Refinement:}
Mask points lift a 2D detection into 3D better than frustum points~\cite{mccraith2021lifting}, and L shape fitting places vehicle boxes on laser scans~\cite{zhang2017efficient}.
Misalignment and scan timing leave background points in masks and offsets in GT boxes~\cite{li2025look,miro2026correcting}.
The closest temporal work registers box sequences to the point cloud stream by differentiable optimization on synthetic data~\cite{li2026differentiable}.
MotionSync refines the output of a causal tracker with a non causal smoothing pass over buffered tracks~\cite{ahuja2026motionsync}.
Both start from LiDAR boxes, while \hestia lifts SAM masks and corrects each box with a \refiner before \fusion.
The \refiner follows the 2D to 3D box regression of Frustum PointNets~\cite{qi2018frustum}, a PointNet~\cite{qi2016pointnet} on the points behind a 2D detection.
It belongs to the lineage of learned 3D annotators that regress boxes from 2D boxes and LiDAR points~\cite{liu2022map,liu2022multimodal,qian2023context}.
Unlike them, our \refiner starts from a mask cluster with a geometric proposal, and its training injects noise into both.

\paragraph{Heading Estimation:}
Image heads estimate orientation with angle bins, angle distributions or flip awareness~\cite{mousavian20173d,prokudin2018deep,cui2020uncertainty}.
AB3DMOT flips a box that opposes its track, and MonoSOWA resolves flips from motion without human labels~\cite{weng2020ab3dmot,skvrna2025monosowa}.
The \yawmodel of \hestia learns full circle heading from LiDAR clusters and image crops.

\paragraph{Datasets for Dense Traffic:}
Indian and Bangladeshi datasets record dense traffic~\cite{dokania2022idd,parikh2024idd,zunair2024rsud20k}, and \hestia fine tunes \armB on RSUD20K and IDD.

\section{The HESTIA Labeler}
\label{sec:method}

\hestia turns synchronized camera images and LiDAR sweeps into 3D boxes with a class, a heading and a track identity (\cref{fig:pipeline}).
The image stages decide which objects exist and what they are (\cref{sec:method_image}).
Their masks are lifted into the LiDAR (\cref{sec:method_lift}), and a \refiner corrects each box (\cref{sec:method_refiner}).
The \fusion follows, and the \yawmodel then sets the heading before the export (\cref{sec:method_fusion}).
\cref{sec:method_finetune} covers the local classes.

\subsection{Image Stages}
\label{sec:method_image}

\paragraph{Detection and Merge:}
\label{sec:method_detect}
Two 2D detectors see every image, and we call them \armA and \armB.
\ArmA is a YOLO11x detector with COCO weights that covers the common classes~\cite{jocher2024ultralytics,lin2014microsoft}.
It is fine tuned on H3D~\cite{ettinger2021h3d}, KITTI~\cite{geiger2012kitti}, Waymo Open~\cite{sun2019scalability}, BDD100K~\cite{yu2018bdd100k} and Argoverse 2~\cite{wilson2023argoverse}.
The same architecture fine tuned on RSUD20K and IDD~\cite{zunair2024rsud20k,varma2019idd} gives \armB, which reports cycle rickshaws and auto rickshaws.
Both arms keep the detections above a confidence score floor and give each one a class name as text.
The names are the ten nuScenes detection classes plus the two local classes.
When two boxes overlap, a class authority table decides which one survives.
This fixed table names the detector that is trusted for each class.
The detector score plays no part in that choice.
Cars, trucks, buses, motorcycles and bicycles of \armA lose to \armB, since \armA has no rickshaw class and names a rickshaw after the nearest vehicle.

\paragraph{Bearing Gate:}
\label{sec:method_gate}
A vehicle cut off by the image edge has an unreliable 2D box.
The \gate does not drop such a box, and SAM tracking carries the object from frames that show it in full.
The bearing \bearing{} of the camera ray through the box center is measured against the viewing direction \(\phi_c\) of that camera.
A box keeps \emph{detect} mode when
\begin{equation}
  \bigl|\operatorname{wrap}(\bearing - \phi_c)\bigr| \le \tau_c,
  \qquad
  \tau_c = \arctan\!\frac{W_c}{2 f_{x,c}} - \delta ,
  \label{eq:gate}
\end{equation}
where \(W_c\) is the image width, \(f_{x,c}\) the focal length in pixels and \(\delta\) a fixed margin.
Outside this angular sector a box is in \emph{track only} mode and does not start, match or relabel a track.
The two Dhaka sessions of \cref{sec:results} use only the front and back cameras, whose axes lie at \ang{0} and \ang{180}.
Boxes whose confidence score lies between the score floor and a higher main floor form the \emph{bypass band}.
Such a bypass box in detect mode may join an existing track at matching.
Otherwise the \verifier judges it alone, provided its crop is large enough.

\paragraph{Tracking:}
\label{sec:method_track}
SAM~3.1 tracks each detect mode box through the video with its multiplex video predictor~\cite{carion2025sam3,meta2026sam31}.
The box is given to SAM~3.1 as two labeled corner points.
A Hungarian assignment joins detections to tracks at every keyframe.
The affinity of a detection and a track adds a term from the LiDAR points and a term from the 2D boxes.
LiDAR thus helps to associate objects and never proposes one.
The tracks of the single cameras are then stitched.
Two tracks are linked when they show one object at the same time or one after the other, and linked tracks share one global identity.

\paragraph{VLM Verification:}
\label{sec:method_vlm}
The \verifier checks the class of every track and every unmatched bypass box with a blind prompt that names no detector label and no city.
Each query shows one crop slightly larger than the box, and the answer is a class name, \etype{not an object} or \etype{unclear}.
For a track the query uses the best crop above a minimum crop size, the crop floor.
The val subset runs ask once per track, and the full runs take a vote over three such crops.
Without such a crop, a track keeps its detector label for review.
A \emph{confirmed} verdict keeps the class, a \emph{relabeled} verdict replaces it and \etype{not an object} \emph{rejects} it.
An unclear or invalid answer keeps the detector class with a review flag.
A bypass box shown to the \verifier survives only when confirmed or relabeled. A smaller bypass box skips it, and the export lifts it at half score.
Each class name carries a typical size, which the stitching and the exported record use as a prior.

\subsection{Mask Sources and Lift}
\label{sec:method_lift}

\paragraph{Prompt Arm:}
The prompt arm gives SAM~3.1 one text prompt per nuScenes class and adds masks for classes that \armA lacks or finds poorly.

\paragraph{Lift and Proposal:}
LiDAR points are projected into the camera, with the motion of the vehicle between the two sensor times compensated.
A depth test keeps the visible points inside the mask, which is first shrunk at its border.
All masks of one global identity then merge into one cluster of points.
Random sample consensus (RANSAC) removes the ground~\cite{fischler1981random}, and region growing collects the object~\cite{wei2021fgr}.
An L shape fit then gives a first box, the fitted box~\cite{zhang2017efficient}.
The fitted box is fused with the size prior, extended away from the sensor and anchored to the ground.
The result is the proposal box, whose heading is known only modulo \ang{180}.

\subsection{Box Refiner and Heading}
\label{sec:method_refiner}

\paragraph{Refiner:}
The \refiner is a PointNet, a network on unordered points~\cite{qi2016pointnet}.
It corrects the proposal box from the points in the coordinate frame of that box.
Its heading output picks one of four candidates at multiples of \ang{90} and adds a residual within \ang{45}.
Further outputs estimate the quality of the output box and the uncertainty of its parameters.
Quality means the expected overlap of that box with the true box.
Training uses labels and camera panoptic masks of Waymo Open~\cite{sun2019scalability}, with seeded noise in masks, LiDAR, size priors and the proposal box.
Model selection uses held out train segments.

\paragraph{Yaw Model:}
\label{sec:method_yaw}
The \yawmodel estimates the heading \yaw{} of each image stage object per keyframe.
It runs after the \fusion, reads the tracks of that step and estimates the axis of the object and its direction along that axis separately.
An image head estimates the observation angle on a ResNet-50 image backbone~\cite{he2015deep}.
It uses MultiBin, a binned angle head~\cite{mousavian20173d}.
A fixed rule then averages the axis from the image with the axes from the LiDAR points, each with a weight.
A bidirectional gated recurrent unit (GRU) then reads the whole track and outputs the heading, the probability that it is reversed and a direction confidence.
Frames with a low direction confidence go to review.
All parts of the \yawmodel learn from Waymo Open labels.

\subsection{Temporal Fusion, Export and Review}
\label{sec:method_fusion}

The \fusion joins the refined boxes of all keyframes into tracks with a greedy constant velocity tracker.
Each track keeps one size, the inverse variance weighted mean of its boxes.
A Kalman filter~\cite{kalman1960new} and a Rauch Tung Striebel smoother~\cite{rauch1965maximum} smooth its centers.
Both are weighted by the uncertainty that the \refiner predicts.
Within the \fusion, moving tracks take their heading from the smoothed motion~\cite{ahuja2026motionsync} and all other tracks keep the \refiner heading.
The export writes a nuScenes detection file, reads no GT and scores each box as
\begin{equation}
  s = s_{\mathrm{det}} \, q^{\gamma},
  \label{eq:score}
\end{equation}
where \(s_{\mathrm{det}}\) is the detector score, \(q\) the output quality of the \refiner and \(\gamma\) a fixed exponent.
For a prompt arm box \(s_{\mathrm{det}}\) is the SAM~3.1 mask score.
Velocity comes from the smoother, and boxes that a registration of the LiDAR sweeps finds static get zero velocity, except pedestrians.
The export also lifts the bypass boxes that were too small for the \verifier, at half their detector score.
Where two cameras overlap, clusters of the same class that share LiDAR points merge into one object.
The \yawmodel then overrides both headings for every box of the image stages, except barriers and traffic cones.
Prompt arm boxes and boxes below the crop floor keep the heading of the \fusion.
Every box keeps a record of its source detector, its \verifier verdict and its review flags.
The record serves an opt in review in the CVAT annotation tool~\cite{cvat2026cvat}.

\subsection{Local Classes}
\label{sec:method_finetune}

Three parts carry the local classes of South Asian traffic.
\ArmB is trained on RSUD20K and IDD and proposes cycle rickshaws and auto rickshaws.
The \verifier carries a low rank adapter trained on crops of driving and South Asian datasets under its blind prompt.
On Dhaka the \refiner has a class slot and a size prior for each rickshaw class, without training on either.
SAM~3.1 runs with its released weights and carries no local class.

\section{Setup}
\label{sec:setup}

\paragraph{Data:}
We run \hestia on the \TrainLabelScenes{} train scenes and the \FullValScenes{} val scenes of nuScenes \cite{caesar2020nuscenes}, with six cameras per keyframe.
The val split of nuScenes holds \FullValKeyframes{} keyframes.
No weight was trained on the val scenes.

\paragraph{Protocols:}
The first of four protocols is the evaluator of \vespa \cite{tempfli2026vespa} on the GT free export of \hestia.
This evaluator maps the classes to one, three or eight classes and drops barrier and traffic cone.
The eight are car, truck, bus, trailer, construction vehicle, pedestrian, motorcycle and bicycle, and the three are vehicle, pedestrian and bicycle.
Its one class mapping counts pedestrians and cyclists as vehicles.
The second protocol is the nuScenes devkit, which scores the same export with the nuScenes detection score (\nds) and the mean orientation error (mAOE).
Both evaluators match boxes to GT by center distance, so a new heading moves \nds through the orientation error and leaves \map unchanged.
The third protocol is the downstream test of \vespa.
The CenterPoint LiDAR detector~\cite{yin2021center} trains on the pseudo labels of the \TrainLabelScenes{} train scenes above a confidence score floor.
It is scored on the \FullValScenes{} val scenes.
The floor maximizes the F1 score of the pseudo labels against the train GT.
In the fourth protocol the nuScenes devkit scores the labels as tracks of seven classes with the average multi object tracking accuracy (AMOTA).
It counts identity switches and track fragments as well.

\begin{figure*}[t]
\centering
\IfFileExists{figs/fig_grid.pdf}{\includegraphics[width=\textwidth]{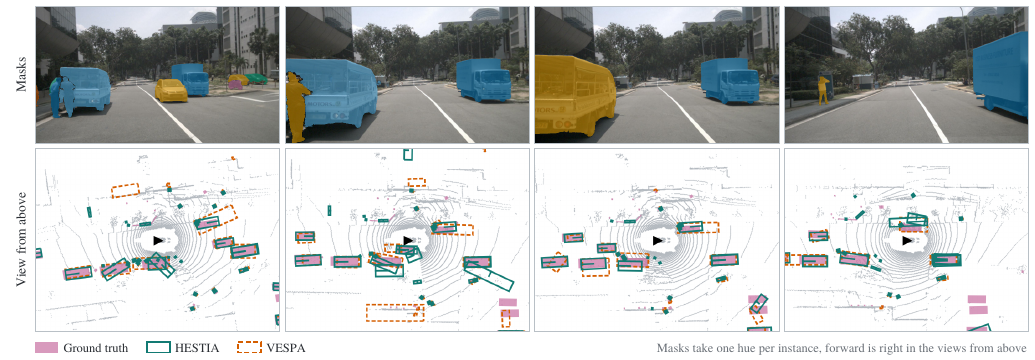}}{\phfig{3.5in}{fig\_grid}}
\caption{\textbf{Four keyframes of scene-0966:} Each column is one keyframe.
The top row shows the front camera with the \hestia masks, one hue per instance.
The bottom row shows the LiDAR points from above within \SI{30}{m} ahead and behind.
GT boxes are filled pink, \hestia pseudo labels are solid outlines with a heading tick and \vespa pseudo labels are dashed outlines.
Boxes are drawn after the evaluator's range and score filters.
The \hestia boxes come from the val subset run with an earlier \refiner and \yawmodel.}
\label{fig:qual}
\end{figure*}

\afterpage{%
\begin{table*}[t]
\centering
\footnotesize
\setlength{\tabcolsep}{5pt}
\caption{\textbf{The VESPA protocol on nuScenes:} Pseudo labels are scored on the \TrainLabelScenes{} train scenes.
CenterPoint detectors trained on each label set are scored on the \FullValScenes{} val scenes.
The rows above \hestia are quoted from the \vespa paper, except the supervised row, which is our own run.
Every pseudo label box of every row has a confidence score of one.
CM3D includes self training, and a dash marks a value that is not reported.
Bold marks the best pseudo label row of each column.}
\label{tab:protocol}
\begin{tabular*}{\textwidth}{@{\extracolsep{\fill}}l cccccccccc@{}}
\toprule
& \multicolumn{4}{c}{Pseudo labels on \TrainLabelScenes{} train scenes} & \multicolumn{6}{c}{CenterPoint on \FullValScenes{} val scenes} \\
\cmidrule(r{7pt}){2-5}\cmidrule(l){6-11}
& \multicolumn{2}{c}{3 classes} & \multicolumn{2}{c}{8 classes} & \multicolumn{2}{c}{1 class} & \multicolumn{2}{c}{3 classes} & \multicolumn{2}{c}{8 classes} \\
\cmidrule(lr){2-3}\cmidrule(r{7pt}){4-5}\cmidrule(lr){6-7}\cmidrule(lr){8-9}\cmidrule(l){10-11}
Labels & mAP\,$\uparrow$ & NDS\,$\uparrow$ & mAP\,$\uparrow$ & NDS\,$\uparrow$ & mAP\,$\uparrow$ & NDS\,$\uparrow$ & mAP\,$\uparrow$ & NDS\,$\uparrow$ & mAP\,$\uparrow$ & NDS\,$\uparrow$ \\
\midrule
Supervised, GT, our run & -- & -- & -- & -- & \DetGTOnemAP & \DetGTOneNDS & \DetGTThreemAP & \DetGTThreeNDS & \DetGTEightmAP & \DetGTEightNDS \\
\addlinespace[1.5pt]
UNION~\cite{lentsch2024union} & \BestQuotedUnionPLThreemAP & \BestQuotedUnionPLThreeNDS & \BestQuotedUnionPLEightmAP & \BestQuotedUnionPLEightNDS & \BestQuotedUnionCPOnemAP & \BestQuotedUnionCPOneNDS & \BestQuotedUnionCPThreemAP & \BestQuotedUnionCPThreeNDS & \BestQuotedUnionCPEightmAP & \BestQuotedUnionCPEightNDS \\
CM3D~\cite{khurana2024cm3d} & -- & -- & -- & -- & \BestQuotedCMCPOnemAP & \BestQuotedCMCPOneNDS & -- & -- & -- & -- \\
AnnofreeOD~\cite{sun2025annofreeod} & -- & -- & -- & -- & \BestQuotedAnnoFreeCPOnemAP & \BestQuotedAnnoFreeCPOneNDS & \BestQuotedAnnoFreeCPThreemAP & \BestQuotedAnnoFreeCPThreeNDS & \BestQuotedAnnoFreeCPEightmAP & \BestQuotedAnnoFreeCPEightNDS \\
\vespa~\cite{tempfli2026vespa} & \BestQuotedVespaPLThreemAP & \BestQuotedVespaPLThreeNDS & \BestQuotedVespaPLEightmAP & \BestQuotedVespaPLEightNDS & \BestQuotedVespaCPOnemAP & \BestQuotedVespaCPOneNDS & \BestQuotedVespaCPThreemAP & \BestQuotedVespaCPThreeNDS & \BestQuotedVespaCPEightmAP & \BestQuotedVespaCPEightNDS \\
\hestia & \BestTrainLabelHestiaVtwoThreemAP & \BestTrainLabelHestiaVtwoThreeNDS & \BestTrainLabelHestiaVtwoEightmAP & \BestTrainLabelHestiaVtwoEightNDS & \BestDetHestiaVtwoOnemAP & \BestDetHestiaVtwoOneNDS & \BestDetHestiaVtwoThreemAP & \BestDetHestiaVtwoThreeNDS & \BestDetHestiaVtwoEightmAP & \BestDetHestiaVtwoEightNDS \\
\bottomrule
\end{tabular*}
\end{table*}}

\paragraph{Baselines:}
\vespa \cite{tempfli2026vespa} is our baseline, and we quote its pseudo label and detector numbers on nuScenes from its paper (\cref{tab:protocol}).
Every \vespa box carries a confidence score of one, and so does every \hestia box.
Neither label \map ranks its boxes.
In use a user keeps the boxes they trust and drops the rest.
UNION~\cite{lentsch2024union}, CM3D~\cite{khurana2024cm3d} and AnnofreeOD~\cite{sun2025annofreeod} are quoted from the same paper.
The supervised detector is our own run of the CenterPoint recipe on the train GT.
On Dhaka data and in \cref{fig:qual} \vespa runs with its public code, its own 2D detector and its default thresholds.
On Dhaka two things change.
The local class names enter as extra prompts with authored size priors, and the ego origin moves down to the road.
We also reran \vespa on the val subset, which gives the boxes of \cref{fig:qual}.
Pseudo label quality is the main endpoint, and the downstream detector of \cref{tab:protocol} tests it.

\paragraph{DHAKA200:}
\dhakadata is a sample of \DkAllScenes{} scenes of dense Dhaka traffic that we recorded and store in the nuScenes format.
Our vehicle carries a Mid-360 LiDAR, six cameras, an IMU and GNSS.
A scene lasts up to \DkSceneSec{}\,\si{s} with keyframes at \DkKeyframeHz{}\,\si{Hz}, and the nuScenes devkit reads its tables unchanged.
Its dense mixed traffic includes the local classes cycle rickshaw and auto rickshaw.
The split holds \DkTrainScenes{} train scenes with \DkTrainKeyframes{} keyframes and \DkValScenes{} val scenes.
Its \DkTestScenes{} test scenes hold \DkTestKeyframes{} keyframes.
The Dhaka benchmark, the Dhaka detectors and the label release of this paper use these scenes.
\dhakadata and its \hestia labels are released under CC~BY-NC~4.0.
\hestia labels the train scenes with the same \armA and with \armB.
\vespa labels them with the Dhaka changes named above.
CenterPoint trains on each label set in the one, three and eight class mappings, as in \cref{tab:protocol}.
No score floor is applied here.
On \dhakadata the eight classes are car, truck, bus, pedestrian, motorcycle, bicycle, cycle rickshaw and auto rickshaw.
The test scenes carry hand drawn 2D boxes that the \hestia 3D stages lift into LiDAR (\cref{sec:results}).
Both detectors are scored against these lifted boxes.

\paragraph{Models and Training:}
On nuScenes \armA is YOLO11x \cite{jocher2024ultralytics}, which starts from COCO weights \cite{lin2014microsoft}.
It is fine tuned on H3D, KITTI, Waymo Open, BDD100K and Argoverse 2, and \armB is off there.
Those datasets carry human 2D boxes, so these runs are free of 3D labels at label time but not of image labels.
The \verifier adapter was trained on crops of driving and South Asian datasets.
The adapter chose its checkpoint on held out crops of the same datasets, and the \verifier votes over three frames.
Masks and tracks come from SAM~3.1 \cite{carion2025sam3,meta2026sam31}.
The \verifier is Nemotron 3 Nano Omni 30B A3B~\cite{nvidia2026nemotron}, an openly released vision language model.
It is served in four bit floating point (NVFP4) on the vLLM engine \cite{kwon2023efficient}.
The \refiner and the \yawmodel are our own models, trained in house with the recipes of \cref{sec:method_refiner}.
For the nuScenes results we use the \refiner and the \yawmodel trained only on Waymo Open, so no nuScenes label enters the training of either.
The size priors of the lift remain the class means of nuScenes train, the split on which the pseudo labels are scored.
An earlier \refiner and \yawmodel trained on nuScenes train labels serve the val subset studies and the figures.
The Dhaka labels also come from a \refiner and a \yawmodel trained on nuScenes train labels.

\paragraph{Hardware:}
For the val subset runs, the 2D stages ran on a GB10 workstation and an RTX 4090.
The full run served the \verifier from the GB10.
Its 2D stages ran on an RTX PRO 6000, an RTX 4090 and two RTX 4080 Super.

\section{Results}
\label{sec:results}

\paragraph{Pseudo Labels:}
On the \TrainLabelScenes{} train scenes the \TrainLabelHestiaBoxes{} \hestia pseudo labels reach \TrainLabelHestiaVtwoEightmAP{} eight class \map against the train GT (\cref{tab:protocol}).
\vespa reports \QuotedVespaPLEightmAP{} there with every score set to one, in a comparison across papers.
In the three class mapping the \hestia pseudo labels reach \TrainLabelHestiaVtwoThreemAP{} against \QuotedVespaPLThreemAP{} for \vespa.
\cref{fig:qual} overlays GT, \hestia and \vespa boxes on four keyframes of one val scene.
Missed objects and boxes without GT can be read off there.

\paragraph{Ablations:}
Our \refiner and \yawmodel are trained on Waymo Open and then label nuScenes, a dataset they never train on.
\cref{tab:learned} gives the gain of each model on nuScenes val.
On \WmNusRefN{} boxes matched to GT the \refiner raises \iou from \WmNusRefInputIou{} to \WmNusRefWIou{}.
Its median heading error modulo \ang{180} rises from \WmNusRefInputYaw{} to \WmNusRefWYaw{} degrees.
On all val scenes our \yawmodel raises devkit \nds from \FullHestiaVtwoDevNDSPreYaw{} to \FullHestiaVtwoDevNDS{}.
The \yawmodel lowers the mean orientation error from \FullHestiaVtwoDevmAOEPreYaw{} to \FullHestiaVtwoDevmAOE{} radians.
On the val scenes the smoothing of the \fusion raises AMOTA from \TvNoFusAMOTA{} to \TvFusAMOTA{} and cuts track fragments from \TvNoFusFRAG{} to \TvFusFRAG{} (\cref{tab:tracking}).
Bus AMOTA rises from \TvNoFusAMOTABus{} to \TvFusAMOTABus{} and truck AMOTA from \TvNoFusAMOTATruck{} to \TvFusAMOTATruck{}.
Identity switches rise from \TvNoFusIDS{} to \TvFusIDS{} because the fused boxes match more GT objects.
Per matched box the switch rate is \TvNoFusIDSRate{} percent without the smoothing and \TvFusIDSRate{} percent with it.
Fused along the stitched SAM~3.1 track ids, the labels reach only \TvSamAMOTA{} AMOTA.
Prompt arm boxes and small bypass boxes carry no such id, so each stays a track of its own.
On the \TrainLabelScenes{} train scenes the smoothing of the \fusion raises AMOTA from \TtNoFusAMOTA{} to \TtFusAMOTA{} and cuts track fragments from \TtNoFusFRAG{} to \TtFusFRAG{}.
Identity switches stay at about \TtFusIDS{} there, and the switch rate falls from \TtNoFusIDSRate{} to \TtFusIDSRate{} percent.

\begin{table}[t]
\centering
\footnotesize
\setlength{\tabcolsep}{1pt}
\caption{\textbf{Tracking on nuScenes:} The nuScenes devkit scores the labels of the \FullValScenes{} val and the \TrainLabelScenes{} train scenes as tracks of seven classes.
Boxes without smoothing keep the greedy tracks of the \fusion and lose only its smoothing.
SAM ids fused links boxes over time by the stitched SAM~3.1 track ids of the image stages instead of the greedy 3D tracker.
IDS counts identity switches, FRAG track fragments and MT mostly tracked objects.
Bold marks the best value of each column within a block.}
\label{tab:tracking}
\begin{tabular*}{\linewidth}{@{\extracolsep{\fill}}lcccccc@{}}
\toprule
Labels & AMOTA $\uparrow$ & MOTA $\uparrow$ & Recall $\uparrow$ & IDS $\downarrow$ & FRAG $\downarrow$ & MT $\uparrow$ \\
\midrule
\multicolumn{7}{@{}l}{\textit{Val, \FullValScenes{} scenes}} \\
With smoothing & \BestTvFusAMOTA & \BestTvFusMOTA & \BestTvFusRecall & \BestTvFusIDS & \BestTvFusFRAG & \BestTvFusMT \\
Without smoothing & \BestTvNoFusAMOTA & \BestTvNoFusMOTA & \BestTvNoFusRecall & \BestTvNoFusIDS & \BestTvNoFusFRAG & \BestTvNoFusMT \\
SAM ids fused & \BestTvSamAMOTA & \BestTvSamMOTA & \BestTvSamRecall & \BestTvSamIDS & \BestTvSamFRAG & \BestTvSamMT \\
\midrule
\multicolumn{7}{@{}l}{\textit{Train, \TrainLabelScenes{} scenes}} \\
With smoothing & \BestTtFusAMOTA & \BestTtFusMOTA & \BestTtFusRecall & \BestTtFusIDS & \BestTtFusFRAG & \BestTtFusMT \\
Without smoothing & \BestTtNoFusAMOTA & \BestTtNoFusMOTA & \BestTtNoFusRecall & \BestTtNoFusIDS & \BestTtNoFusFRAG & \BestTtNoFusMT \\
SAM ids fused & \BestTtSamAMOTA & \BestTtSamMOTA & \BestTtSamRecall & \BestTtSamIDS & \BestTtSamFRAG & \BestTtSamMT \\
\bottomrule
\end{tabular*}
\end{table}

\begin{table}[t]
\centering
\footnotesize
\setlength{\tabcolsep}{4pt}
\caption{\textbf{Box refiner and yaw model on nuScenes val:} Both models are trained only on Waymo Open.
The upper block scores \WmNusRefN{} proposal boxes matched to GT before and after the \refiner.
Center is the median center error and Yaw the median heading error modulo \ang{180}.
The lower block scores the labels of all val scenes under the nuScenes devkit before and after the \yawmodel.
Bold marks the better value of each pair.}
\label{tab:learned}
\begin{tabular}{@{}lcccc@{}}
\toprule
Box & 3D IoU & Acc@0.5 (\si{\percent}) & Center (\si{m}) & Yaw (\si{\degree}) \\
\midrule
Proposal box & \BestWmNusRefInputIou & \BestWmNusRefInputAcc & \BestWmNusRefInputCtr & \BestWmNusRefInputYaw \\
After the \refiner & \BestWmNusRefWIou & \BestWmNusRefWAcc & \BestWmNusRefWCtr & \BestWmNusRefWYaw \\
\midrule
Labels & \multicolumn{2}{c}{NDS} & \multicolumn{2}{c}{mAOE (rad)} \\
\midrule
Before the \yawmodel & \multicolumn{2}{c}{\BestFullHestiaVtwoDevNDSPreYaw} & \multicolumn{2}{c}{\BestFullHestiaVtwoDevmAOEPreYaw} \\
After the \yawmodel & \multicolumn{2}{c}{\BestFullHestiaVtwoDevNDS} & \multicolumn{2}{c}{\BestFullHestiaVtwoDevmAOE} \\
\bottomrule
\end{tabular}
\end{table}
The single factors of the export were ablated on the val subset with the earlier \refiner and \yawmodel.
There the prompt arm adds the most \map.
The earlier \yawmodel lowers mAOE the most, and lifting the small bypass boxes and zeroing the velocity of static boxes raise \nds.

\paragraph{Downstream Detector:}
As in \vespa, \hestia labels all \TrainLabelScenes{} train scenes, and the CenterPoint LiDAR detector trains on these pseudo labels (\cref{tab:protocol}).
The class agnostic CenterPoint detector trains on the \hestia pseudo labels above the confidence score floor of \cref{sec:setup}, chosen against the train GT.
It scores \DetHestiaVtwoOnemAP{} \map and \DetHestiaVtwoOneNDS{} \nds on val.
For CenterPoint trained on its own pseudo labels, \vespa reports \QuotedVespaCPOnemAP{} and \QuotedVespaCPOneNDS{}.
The same CenterPoint recipe trained on GT scores \DetGTOnemAP{}.
The detector on \hestia pseudo labels thus reaches \DetHestiaVtwoOneSharemAP{} percent of the class agnostic \map of full supervision.

\paragraph{Image Stages:}
The image stages were studied on the val subset.
There the \verifier raises 2D \map from \TwoDmAPNoVlm{} to \TwoDmAPVlm{}, mostly by renaming boxes into construction vehicle and trailer.
On RSUD20K test \armB reaches \PAArmBLocal{} AP50 on the local classes, the 2D average precision at IoU 0.5.
The merge of \armA and \armB keeps \PAMergeLocal{} where \armA alone finds none.

\paragraph{Verifier Corrections:}
On the val subset we scored the class changes and rejections of the \verifier against the nuScenes GT.
Of its class changes, \VlmFixedShare{} percent gave a box its GT class and \VlmBrokeShare{} percent replaced a correct class.
Of the boxes it rejected, \VlmRejectedPrec{} percent match no GT object.
After the \verifier, \TwoDTypeAccPct{} percent of the matched boxes carry the GT class.
The study covers \NScenes{} scenes.
Scaled to the \TrainLabelScenes{} train scenes, the \verifier fixed about \VlmFixedTrain{} box classes, broke about \VlmBrokeTrain{} and rejected about \VlmRejectedTrain{} boxes.

\paragraph{Dense Traffic in Dhaka:}
Apart from \dhakadata, \hestia labeled two Dhaka sessions with only the front and the back camera.
Neither session is part of the release.
Session A holds \DhakaKeyframes{} keyframes and session B \DhakaBKeyframes{}, and a 2D row is one shipped box in one image.
A front or back image holds \DhakaRowsPerImage{} 2D rows in session A and \DhakaBRowsPerImage{} in session B, against \NusRowsPerImage{} on the nuScenes val subset.
The traffic is denser in the image only, since the 3D density within the LiDAR disc is close to that of nuScenes in both sessions.
\ArmB proposes \DhakaArmBShare{} percent of the 2D rows in session A.
Both sessions ran an earlier \armB with the same two local classes.
On the same session B keyframes, both sets kept within 50 m, \vespa writes \DVVespaBoxes{} boxes against \DVHestiaBoxes{} for \hestia.
It agrees with \DVVespaLegacyAuto{} percent of the auto rickshaw machine pre labels made for this session before \hestia, and \hestia with \DVHestiaLegacyAuto{}.
\vespa ran with an ego origin moved to the road and authored priors.

\begin{table}[t]
\centering
\footnotesize
\setlength{\tabcolsep}{3pt}
\caption{\textbf{Detectors on DHAKA200:} CenterPoint trained on the \vespa labels and on the \hestia labels of the \DkTrainScenes{} train scenes of \dhakadata.
Both are scored on \DkTestScenes{} test scenes in the one, three and eight class mappings.
The rickshaws join the vehicle class in the three class mapping.
The test objects are hand drawn 2D boxes of those scenes, lifted into LiDAR by the \hestia 3D stages.
Bold marks the higher value of each column.}
\label{tab:dhakasense}
\begin{tabular*}{\linewidth}{@{\extracolsep{\fill}}lcccccc@{}}
\toprule
& \multicolumn{2}{c}{1 class} & \multicolumn{2}{c}{3 classes} & \multicolumn{2}{c}{8 classes} \\
\cmidrule(lr){2-3}\cmidrule(lr){4-5}\cmidrule(l){6-7}
Train labels & mAP & NDS & mAP & NDS & mAP & NDS \\
\midrule
\vespa & \BestDkDetVespaOnemAP & \BestDkDetVespaOneNDS & \BestDkDetVespaThreemAP & \BestDkDetVespaThreeNDS & \BestDkDetVespaEightmAP & \BestDkDetVespaEightNDS \\
\hestia & \BestDkDetHestiaOnemAP & \BestDkDetHestiaOneNDS & \BestDkDetHestiaThreemAP & \BestDkDetHestiaThreeNDS & \BestDkDetHestiaEightmAP & \BestDkDetHestiaEightNDS \\
\bottomrule
\end{tabular*}
\end{table}

\paragraph{DHAKA200 Benchmark:}
We train CenterPoint on the pseudo labels of \hestia and of \vespa and score each detector on the test scenes of \dhakadata (\cref{tab:dhakasense}).
The \hestia train labels hold \DkHestiaTrainBoxes{} boxes and the \vespa labels \DkVespaTrainBoxes{}.
The CenterPoint detector trained on \hestia labels scores \DkDetHestiaEightmAP{} eight class \map, and the one trained on \vespa labels \DkDetVespaEightmAP{}.
In the class agnostic mapping the two detectors score \DkDetHestiaOnemAP{} and \DkDetVespaOnemAP{} \map.
The detector on \hestia labels scores higher there too, in a mapping that counts every class as one, pedestrians included.
On cycle rickshaws the two detectors reach \DkDetHestiaAPCycleRickshaw{} and \DkDetVespaAPCycleRickshaw{} AP, and on auto rickshaws \DkDetHestiaAPAutoRickshaw{} and \DkDetVespaAPAutoRickshaw{}.
The test reference is not independent of \hestia.
Five annotators, all authors, drew and verified 2D boxes in a browser tool.
SAM 3.1 tracked each box, and the \hestia 3D stages lifted it into LiDAR.
Humans thus decide which objects the reference holds, and our own lift sets their 3D geometry.
Both detectors are scored against these lifted boxes, which may favor the detector trained on \hestia labels.
The reference covers objects within about 30 m.

\paragraph{Verifier Adapter:}
On crops of held out \dhakadata scenes the fine tuned adapter raises the accuracy of the \verifier from \LoraVerVanAcc{} to \LoraVerFtAcc{} percent.
The class labels of these crops are machine pre labels.
On crops of the nuScenes val subset the accuracy rises from \LoraNusVanAcc{} to \LoraNusFtAcc{} percent.

\paragraph{Label Release:}
We release the \hestia labels of \dhakadata, which come from the \hestia pipeline alone and which no annotator has verified.
We did not score the released labels against an independent 3D reference.
Their only quality evidence is the detector of \cref{tab:dhakasense}.
Every box records the detector arm that proposed it, the verdict of the \verifier, its review flags and the stages that set its fields.
The SAM~3.1 instance masks ship with the boxes.

\section{Limitations}
\label{sec:limitations}
\paragraph{In House Models Lose Accuracy in HESTIA:}
The \refiner and the \yawmodel are trained in house, and their development accuracy does not carry over to GT free labeling.
On GT matched tracks and crops of nuScenes val the \yawmodel reaches a median heading error of \WmNusYawWFinalMed{} degrees.
Its flip rate there is \WmNusYawWFinalFlip{} percent.
On predicted tracks and boxes of all val scenes it lowers the devkit mean orientation error only from \FullHestiaVtwoDevmAOEPreYaw{} to \FullHestiaVtwoDevmAOE{} radians.
The reason is that tracks break, with \TvFusIDS{} identity switches on val (\cref{tab:tracking}), and crops miss the object.
The \refiner brings GT matched boxes of nuScenes val to \WmNusRefWIou{} \iou.
We report both accuracies so that development accuracy is never read as labeling accuracy.

\paragraph{Dhaka Caveats:}
On Dhaka the whole pipeline runs, the \refiner, \fusion and \yawmodel included.
The fused boxes had implausible sizes, so we ship single frame boxes without velocity or track identity.
The detectors of \cref{tab:dhakasense} train on these boxes, so the velocity term of their \nds is not meaningful.
We report tracking metrics where annotated tracks exist, so the tracking numbers of this paper are on nuScenes.
The Dhaka temporal results will be released with the dataset once all 200 scenes are annotated.
The Dhaka calibration is a little off compared with nuScenes, and the cameras have no hardware sync.
The Dhaka test objects cover \DkTestKeyframes{} keyframes of \DkTestScenes{} test scenes with hand drawn 2D boxes.

\paragraph{Confounds of the Headline Numbers:}
The comparison is not label free, because the \refiner and the \yawmodel train on Waymo Open 3D labels.
The common class detector \armA is fine tuned on H3D, KITTI, Waymo Open, BDD100K and Argoverse 2.
The \verifier adapter is trained on labeled crops of driving and South Asian datasets.
Size priors of the lift come from the class means of nuScenes train.
The \verifier has no measured share of the 3D \map, as it is scored only in 2D and prompt arm masks skip it.
The Dhaka test objects are hand drawn 2D boxes of \DkTestScenes{} test scenes, lifted into LiDAR by the \hestia 3D stages.
Both detectors are scored against them, so this reference is not independent of \hestia.
On the val subset the review flags hold \ReviewFlagCatch{} percent of the false positives, and no annotator worked that queue.
\section{Conclusion}
\label{sec:conclusion}

We presented the camera first labeler \hestia. Its 3D chain lifts masks into LiDAR, corrects them with a learned \refiner, fuses them over time and sets the heading with a \yawmodel. On nuScenes val its smoothing raises AMOTA from \TvNoFusAMOTA{} to \TvFusAMOTA{}. We have no tracking result on Dhaka, where the labels are single frame boxes. On the nuScenes train split its pseudo labels reach \TrainLabelHestiaVtwoEightmAP{} eight class \map, where \vespa reports \QuotedVespaPLEightmAP{} with every box scored one. A class agnostic CenterPoint detector trained on these pseudo labels scores \DetHestiaVtwoOnemAP{} \map, above the number \vespa reports, in a comparison across papers. The \refiner and the \yawmodel are trained on Waymo Open labels, so this comparison is not label free. On \dhakadata, detectors trained on \hestia labels and on \vespa labels are scored against hand drawn 2D boxes of the test scenes. The \hestia 3D stages lift them into LiDAR.

\section*{Ethics and Data Statement}
nuScenes~\cite{caesar2020nuscenes} and Waymo Open~\cite{sun2019scalability} are used under their non commercial terms. The earlier \refiner and \yawmodel of the val subset studies and figures encode nuScenes train labels and nuInsSeg masks~\cite{li2022lwsis}. The \yawmodel image head starts from ImageNet weights~\cite{deng2009imagenet}. YOLO11 is under AGPL-3.0 with COCO weights~\cite{lin2014microsoft}, SAM~3.1 under the SAM License and the \verifier under the NVIDIA Open Model Agreement. The \vespa code has no license file. RSUD20K is under CC~BY-NC~4.0~\cite{zunair2024rsud20k}, so \armB labels stay non commercial, and the IDD registration terms apply too. \dhakadata and its labels are released under CC~BY-NC~4.0. An institutional review board approved the study. Faces and license plates are blurred in the released images. Recording did not select or exclude people by race, age or gender. Fine tuning crops stay unreleased, and other crops only query the \mbox{\verifier}.
{
    \small
    \bibliographystyle{ieeenat_fullname}
    \bibliography{refs}

\begin{thebibliography}{64}
\providecommand{\natexlab}[1]{#1}
\providecommand{\url}[1]{\texttt{#1}}
\expandafter\ifx\csname urlstyle\endcsname\relax
  \providecommand{\doi}[1]{doi: #1}\else
  \providecommand{\doi}{doi: \begingroup \urlstyle{rm}\Url}\fi

\bibitem[Ahuja et~al.(2026)Ahuja, Sudhakar, Gupta, Narayanan, Kumar, and
  Yogamani]{ahuja2026motionsync}
Rahul Ahuja, Bala Murali Manoghar~Sai Sudhakar, Shashwata Gupta, Venkatraman
  Narayanan, Varun~Ravi Kumar, and Senthil Yogamani.
\newblock {MotionSync}: Non-causal refinement of causal tracker for
  label-efficient {3D} perception, 2026.
\newblock arXiv:2608.29567.

\bibitem[Caesar et~al.(2020)Caesar, Bankiti, Lang, Vora, Liong, Xu, Krishnan,
  Pan, Baldan, and Beijbom]{caesar2020nuscenes}
Holger Caesar, Varun Bankiti, Alex~H. Lang, Sourabh Vora, Venice~Erin Liong,
  Qiang Xu, Anush Krishnan, Yu Pan, Giancarlo Baldan, and Oscar Beijbom.
\newblock {nuScenes}: A multimodal dataset for autonomous driving.
\newblock In \emph{CVPR}, 2020.
\newblock arXiv:1903.11027.

\bibitem[Carion et~al.(2025)Carion, Gustafson, Hu, Debnath, Hu, Suris, Ryali,
  Alwala, Khedr, Huang, Lei, Ma, Guo, Kalla, Marks, Greer, Wang, Sun,
  R{\"a}dle, Afouras, Mavroudi, Xu, Wu, Zhou, Momeni, Hazra, Ding, Vaze,
  Porcher, Li, Li, Kamath, Cheng, Doll{\'a}r, Ravi, Saenko, Zhang, and
  Feichtenhofer]{carion2025sam3}
Nicolas Carion, Laura Gustafson, Yuan-Ting Hu, Shoubhik Debnath, Ronghang Hu,
  Didac Suris, Chaitanya Ryali, Kalyan~Vasudev Alwala, Haitham Khedr, Andrew
  Huang, Jie Lei, Tengyu Ma, Baishan Guo, Arpit Kalla, Markus Marks, Joseph
  Greer, Meng Wang, Peize Sun, Roman R{\"a}dle, Triantafyllos Afouras,
  Effrosyni Mavroudi, Katherine Xu, Tsung-Han Wu, Yu Zhou, Liliane Momeni,
  Rishi Hazra, Shuangrui Ding, Sagar Vaze, Francois Porcher, Feng Li, Siyuan
  Li, Aishwarya Kamath, Ho~Kei Cheng, Piotr Doll{\'a}r, Nikhila Ravi, Kate
  Saenko, Pengchuan Zhang, and Christoph Feichtenhofer.
\newblock {SAM} 3: {Segment Anything} with concepts, 2025.
\newblock arXiv:2511.16719.

\bibitem[Chen et~al.(2026)Chen, Khaitan, and Djuric]{chen2026improving}
Steven Chen, Shivesh Khaitan, and Nemanja Djuric.
\newblock Improving {3D} labeling in self-driving by inferring vehicle
  information using vision language models.
\newblock In \emph{IV}, 2026.
\newblock arXiv:2605.21747.

\bibitem[Cui et~al.(2020)Cui, Chou, Charland, Vallespi-Gonzalez, and
  Djuric]{cui2020uncertainty}
Henggang Cui, Fang-Chieh Chou, Jake Charland, Carlos Vallespi-Gonzalez, and
  Nemanja Djuric.
\newblock Uncertainty-aware vehicle orientation estimation for joint
  detection-prediction models, 2020.
\newblock arXiv:2011.03114.

\bibitem[{CVAT.ai Corporation}(2026)]{cvat2026cvat}
{CVAT.ai Corporation}.
\newblock {CVAT}: Computer vision annotation tool, 2026.
\newblock Open-source software, MIT license.

\bibitem[Deng et~al.(2009)Deng, Dong, Socher, Li, Li, and
  Fei-Fei]{deng2009imagenet}
Jia Deng, Wei Dong, Richard Socher, Li-Jia Li, Kai Li, and Li Fei-Fei.
\newblock {ImageNet}: A large-scale hierarchical image database.
\newblock In \emph{CVPR}, pages 248--255, 2009.

\bibitem[Deniz and Ko{\c c}yi{\u g}it(2026)]{deniz2026open}
{\"O}mer~Faruk Deniz and Mustafa~Taha Ko{\c c}yi{\u g}it.
\newblock Open-vocabulary {3D} object detection with promptable segmentation,
  2026.
\newblock arXiv:2609.19358.

\bibitem[Dokania et~al.(2023)Dokania, Hafez, Subramanian, Chandraker, and
  Jawahar]{dokania2022idd}
Shubham Dokania, A.~H.~Abdul Hafez, Anbumani Subramanian, Manmohan Chandraker,
  and C.~V. Jawahar.
\newblock {IDD}-{3D}: {Indian} driving dataset for {3D} unstructured road
  scenes.
\newblock In \emph{WACV}, pages 4471--4480, 2023.
\newblock arXiv:2210.12878.

\bibitem[Fan et~al.(2023)Fan, Yang, Mao, Wang, Chen, Wang, and
  Zhang]{fan2023once}
Lue Fan, Yuxue Yang, Yiming Mao, Feng Wang, Yuntao Chen, Naiyan Wang, and
  Zhaoxiang Zhang.
\newblock Once detected, never lost: Surpassing human performance in offline
  {LiDAR} based {3D} object detection.
\newblock In \emph{ICCV}, pages 19763--19772, 2023.

\bibitem[Fischler and Bolles(1981)]{fischler1981random}
Martin~A. Fischler and Robert~C. Bolles.
\newblock Random sample consensus: A paradigm for model fitting with
  applications to image analysis and automated cartography.
\newblock \emph{Communications of the ACM}, 24\penalty0 (6):\penalty0 381--395,
  1981.

\bibitem[Geiger et~al.(2012)Geiger, Lenz, and Urtasun]{geiger2012kitti}
Andreas Geiger, Philip Lenz, and Raquel Urtasun.
\newblock Are we ready for autonomous driving? the {KITTI} vision benchmark
  suite.
\newblock In \emph{CVPR}, pages 3354--3361, 2012.

\bibitem[Ghilotti et~al.(2026)Ghilotti, Brucker, Saidy, Matteucci, Bijelic, and
  Heide]{ghilotti2026unilips}
Filippo Ghilotti, Samuel Brucker, Nahku Saidy, Matteo Matteucci, Mario Bijelic,
  and Felix Heide.
\newblock {UniLiPs}: Unified {LiDAR} pseudo-labeling with geometry-grounded
  dynamic scene decomposition.
\newblock In \emph{3DV}, 2026.
\newblock arXiv:2601.05105.

\bibitem[He et~al.(2016)He, Zhang, Ren, and Sun]{he2015deep}
Kaiming He, Xiangyu Zhang, Shaoqing Ren, and Jian Sun.
\newblock Deep residual learning for image recognition.
\newblock In \emph{CVPR}, pages 770--778, 2016.
\newblock arXiv:1512.03385.

\bibitem[Jiang et~al.(2026)Jiang, Wang, Zhao, Li, and Jiang]{jiang2026sam2mot}
Junjie Jiang, Zelin Wang, Manqi Zhao, Yin Li, and Dongsheng Jiang.
\newblock {SAM2MOT}: A novel paradigm of multi-object tracking by segmentation.
\newblock In \emph{AAAI}, pages 5388--5396, 2026.

\bibitem[Jocher and Qiu(2024)]{jocher2024ultralytics}
Glenn Jocher and Jing Qiu.
\newblock {Ultralytics YOLO11}, 2024.
\newblock Software, version 11.0.0, AGPL-3.0.

\bibitem[Kalman(1960)]{kalman1960new}
R.~E. Kalman.
\newblock A new approach to linear filtering and prediction problems.
\newblock \emph{Journal of Basic Engineering}, 82\penalty0 (1):\penalty0
  35--45, 1960.

\bibitem[Khurana et~al.(2024)Khurana, Peri, Hays, and Ramanan]{khurana2024cm3d}
Mehar Khurana, Neehar Peri, James Hays, and Deva Ramanan.
\newblock Shelf-supervised cross-modal pre-training for {3D} object detection.
\newblock In \emph{CoRL}, 2024.

\bibitem[Kim et~al.(2024)Kim, Ku, Kim, Cha, and Baek]{kim2024vlm}
Junsu Kim, Yunhoe Ku, Jihyeon Kim, Junuk Cha, and Seungryul Baek.
\newblock {VLM}-{PL}: Advanced pseudo labeling approach for class incremental
  object detection via vision-language model.
\newblock In \emph{CVPR Workshops}, pages 4170--4181, 2024.
\newblock arXiv:2403.05346.

\bibitem[Kirillov et~al.(2023)Kirillov, Mintun, Ravi, Mao, Rolland, Gustafson,
  Xiao, Whitehead, Berg, Lo, Doll{\'a}r, and Girshick]{kirillov2023segment}
Alexander Kirillov, Eric Mintun, Nikhila Ravi, Hanzi Mao, Chloe Rolland, Laura
  Gustafson, Tete Xiao, Spencer Whitehead, Alexander~C. Berg, Wan-Yen Lo, Piotr
  Doll{\'a}r, and Ross Girshick.
\newblock {Segment Anything}.
\newblock In \emph{ICCV}, pages 3992--4003, 2023.

\bibitem[Kwon et~al.(2023)Kwon, Li, Zhuang, Sheng, Zheng, Yu, Gonzalez, Zhang,
  and Stoica]{kwon2023efficient}
Woosuk Kwon, Zhuohan Li, Siyuan Zhuang, Ying Sheng, Lianmin Zheng, Cody~Hao Yu,
  Joseph~E. Gonzalez, Hao Zhang, and Ion Stoica.
\newblock Efficient memory management for large language model serving with
  {PagedAttention}.
\newblock In \emph{SOSP}, pages 611--626, 2023.
\newblock arXiv:2309.06180.

\bibitem[Lee et~al.(2025)Lee, Kim, Ryu, Musacchio, and Park]{lee2025openbox}
In-Jae Lee, Mungyeom Kim, Kwonyoung Ryu, Pierre Musacchio, and Jaesik Park.
\newblock {OpenBox}: Annotate any bounding boxes in {3D}.
\newblock In \emph{NeurIPS}, 2025.
\newblock arXiv:2512.01352.

\bibitem[Lentsch et~al.(2024)Lentsch, Caesar, and Gavrila]{lentsch2024union}
Ted Lentsch, Holger Caesar, and Dariu~M. Gavrila.
\newblock {UNION}: Unsupervised {3D} object detection using object
  appearance-based pseudo-classes.
\newblock In \emph{NeurIPS}, 2024.

\bibitem[Li et~al.(2024)Li, Sun, Wang, Xie, Feng, Zhang, Yuan, Xu, Liu, and
  Luo]{li2024segment}
Jianhao Li, Tianyu Sun, Zhongdao Wang, Enze Xie, Bailan Feng, Hongbo Zhang, Ze
  Yuan, Ke Xu, Jiaheng Liu, and Ping Luo.
\newblock Segment, lift and fit: Automatic {3D} shape labeling from {2D}
  prompts.
\newblock In \emph{ECCV}, 2024.
\newblock arXiv:2407.11382.

\bibitem[Li et~al.(2023)Li, Yin, Shi, Li, Yang, and Shen]{li2022lwsis}
Xiang Li, Junbo Yin, Botian Shi, Yikang Li, Ruigang Yang, and Jianbing Shen.
\newblock {LWSIS}: {LiDAR}-guided weakly supervised instance segmentation for
  autonomous driving.
\newblock In \emph{AAAI}, pages 1433--1441, 2023.
\newblock arXiv:2212.03504.

\bibitem[Li et~al.(2025)Li, Hu, Xu, and Kong]{li2025look}
Xiang Li, Zhangchi Hu, Xiao Xu, and Bin Kong.
\newblock Look before you fuse: {2D}-guided cross-modal alignment for robust
  {3D} detection, 2025.
\newblock arXiv:2507.16861.

\bibitem[Li et~al.(2026)Li, Tong, and Hong]{li2026differentiable}
Xuesong Li, Jinguang Tong, and Jie Hong.
\newblock A differentiable optimization framework for registering sequential
  bounding boxes with point cloud stream, 2026.
\newblock arXiv:2609.34103.

\bibitem[Liang et~al.(2024)Liang, Su, Schulter, Garg, Zhao, Wu, and
  Chandraker]{liang2024aide}
Mingfu Liang, Jong-Chyi Su, Samuel Schulter, Sparsh Garg, Shiyu Zhao, Ying Wu,
  and Manmohan Chandraker.
\newblock {AIDE}: An automatic data engine for object detection in autonomous
  driving.
\newblock In \emph{CVPR}, pages 14695--14706, 2024.
\newblock arXiv:2403.17373.

\bibitem[Lin et~al.(2014)Lin, Maire, Belongie, Bourdev, Girshick, Hays, Perona,
  Ramanan, Zitnick, and Doll{\'a}r]{lin2014microsoft}
Tsung-Yi Lin, Michael Maire, Serge Belongie, Lubomir Bourdev, Ross Girshick,
  James Hays, Pietro Perona, Deva Ramanan, C.~Lawrence Zitnick, and Piotr
  Doll{\'a}r.
\newblock {Microsoft} {COCO}: Common objects in context.
\newblock In \emph{ECCV}, pages 740--755, 2014.
\newblock arXiv:1405.0312.

\bibitem[Liu et~al.(2022{\natexlab{a}})Liu, Qian, Huang, Qi, Lam, Tan, and
  Wong]{liu2022multimodal}
Chang Liu, Xiaoyan Qian, Binxiao Huang, Xiaojuan Qi, Edmund Lam, Siew-Chong
  Tan, and Ngai Wong.
\newblock Multimodal transformer for automatic {3D} annotation and object
  detection.
\newblock In \emph{ECCV}, 2022{\natexlab{a}}.

\bibitem[Liu et~al.(2022{\natexlab{b}})Liu, Qian, Qi, Lam, Tan, and
  Wong]{liu2022map}
Chang Liu, Xiaoyan Qian, Xiaojuan Qi, Edmund~Y. Lam, Siew-Chong Tan, and Ngai
  Wong.
\newblock {MAP-Gen}: An automated {3D}-box annotation flow with multimodal
  attention point generator.
\newblock In \emph{ICPR}, 2022{\natexlab{b}}.

\bibitem[Lu et~al.(2025)Lu, Bian, and Shah]{lu2025clipgrader}
Hong Lu, Yali Bian, and Rahul~C. Shah.
\newblock {ClipGrader}: Leveraging vision-language models for robust label
  quality assessment in object detection, 2025.
\newblock arXiv:2503.02897.

\bibitem[Ma et~al.(2023)Ma, Yang, Zhou, Li, Shi, Liu, Yang, Liu, He, Qiao, Li,
  and Li]{ma2023detzero}
Tao Ma, Xuemeng Yang, Hongbin Zhou, Xin Li, Botian Shi, Junjie Liu, Yuchen
  Yang, Zhizheng Liu, Liang He, Yu Qiao, Yikang Li, and Hongsheng Li.
\newblock {DetZero}: Rethinking offboard {3D} object detection with long-term
  sequential point clouds.
\newblock In \emph{ICCV}, pages 6713--6724, 2023.

\bibitem[Ma et~al.(2026{\natexlab{a}})Ma, Hua, and Kong]{ma2026autoexpert}
Yechi Ma, Wei Hua, and Shu Kong.
\newblock {AutoExpert}: Automating {3D} {LiDAR} annotation from expert-crafted
  guidelines.
\newblock In \emph{NeurIPS}, 2026{\natexlab{a}}.
\newblock arXiv:2506.02914.

\bibitem[Ma et~al.(2026{\natexlab{b}})Ma, Yao, Liu, Xiong, He, and
  Wu]{ma2026efficient}
Yalong Ma, Ziying Yao, Xuan Liu, Zhongxia Xiong, Xiaozheng He, and Xinkai Wu.
\newblock Efficient {3D} object annotation via vision-derived pseudo-{LiDAR}
  and vision language model ({VLM}) validation.
\newblock \emph{Transportation Research Part C: Emerging Technologies},
  182:\penalty0 105429, 2026{\natexlab{b}}.

\bibitem[McCraith et~al.(2022)McCraith, Insafutdinov, Neumann, and
  Vedaldi]{mccraith2021lifting}
Robert McCraith, Eldar Insafutdinov, Lukas Neumann, and Andrea Vedaldi.
\newblock Lifting {2D} object locations to {3D} by discounting {LiDAR} outliers
  across objects and views.
\newblock In \emph{ICRA}, pages 2411--2418, 2022.
\newblock arXiv:2109.07945.

\bibitem[{Meta AI}(2026)]{meta2026sam31}
{Meta AI}.
\newblock {SAM} 3.1 (object multiplex) release notes and model card, 2026.
\newblock Model card facebook/sam3.1.

\bibitem[Miro et~al.(2026)Miro, af~Klinteberg, Timus, Asefaw, Khoche,
  Gustafsson, Mansouri, and Daneshtalab]{miro2026correcting}
Alexandre~Justo Miro, Ludvig af Klinteberg, Bogdan Timus, Aron Asefaw, Ajinkya
  Khoche, Thomas Gustafsson, Sina~Sharif Mansouri, and Masoud Daneshtalab.
\newblock Correcting and quantifying systematic errors in {3D} box annotations
  for autonomous driving.
\newblock In \emph{WACV}, 2026.
\newblock arXiv:2601.14038.

\bibitem[Mousavian et~al.(2017)Mousavian, Anguelov, Flynn, and
  Kosecka]{mousavian20173d}
Arsalan Mousavian, Dragomir Anguelov, John Flynn, and Jana Kosecka.
\newblock {3D} bounding box estimation using deep learning and geometry.
\newblock In \emph{CVPR}, 2017.
\newblock arXiv:1612.00496.

\bibitem[{NVIDIA}(2026)]{nvidia2026nemotron}
{NVIDIA}.
\newblock {Nemotron} {3} {Nano} {Omni}: Efficient and open multimodal
  intelligence, 2026.
\newblock arXiv:2604.24954.

\bibitem[Parikh et~al.(2024)Parikh, Saluja, Jawahar, and
  Sarvadevabhatla]{parikh2024idd}
Chirag Parikh, Rohit Saluja, C.~V. Jawahar, and Ravi~Kiran Sarvadevabhatla.
\newblock {IDD-X}: A multi-view dataset for ego-relative important object
  localization and explanation in dense and unstructured traffic.
\newblock In \emph{ICRA}, pages 14815--14821, 2024.
\newblock arXiv:2404.08561.

\bibitem[Patil et~al.(2019)Patil, Malla, Gang, and Chen]{ettinger2021h3d}
Abhishek Patil, Srikanth Malla, Haiming Gang, and Yi-Ting Chen.
\newblock The {H3D} dataset for full-surround 3d multi-object detection and
  tracking in crowded urban scenes.
\newblock In \emph{ICRA}, pages 9552--9557, 2019.
\newblock arXiv:1903.01568.

\bibitem[Prokudin et~al.(2018)Prokudin, Gehler, and Nowozin]{prokudin2018deep}
Sergey Prokudin, Peter Gehler, and Sebastian Nowozin.
\newblock Deep directional statistics: Pose estimation with uncertainty
  quantification.
\newblock In \emph{ECCV}, pages 542--559, 2018.
\newblock arXiv:1805.03430.

\bibitem[Qi et~al.(2017)Qi, Su, Mo, and Guibas]{qi2016pointnet}
Charles~R. Qi, Hao Su, Kaichun Mo, and Leonidas~J. Guibas.
\newblock {PointNet}: Deep learning on point sets for {3D} classification and
  segmentation.
\newblock In \emph{CVPR}, pages 77--85, 2017.
\newblock arXiv:1612.00593.

\bibitem[Qi et~al.(2018)Qi, Liu, Wu, Su, and Guibas]{qi2018frustum}
Charles~R. Qi, Wei Liu, Chenxia Wu, Hao Su, and Leonidas~J. Guibas.
\newblock Frustum {PointNets} for {3D} object detection from {RGB-D} data.
\newblock In \emph{CVPR}, 2018.

\bibitem[Qi et~al.(2021)Qi, Zhou, Najibi, Sun, Vo, Deng, and
  Anguelov]{qi2021offboard}
Charles~R. Qi, Yin Zhou, Mahyar Najibi, Pei Sun, Khoa Vo, Boyang Deng, and
  Dragomir Anguelov.
\newblock Offboard {3D} object detection from point cloud sequences.
\newblock In \emph{CVPR}, pages 6130--6140, 2021.

\bibitem[Qian et~al.(2023)Qian, Liu, Qi, Tan, Lam, and Wong]{qian2023context}
Xiaoyan Qian, Chang Liu, Xiaojuan Qi, Siew-Chong Tan, Edmund Lam, and Ngai
  Wong.
\newblock Context-aware transformer for {3D} point cloud automatic annotation.
\newblock In \emph{AAAI}, 2023.

\bibitem[Rauch et~al.(1965)Rauch, Tung, and Striebel]{rauch1965maximum}
H.~E. Rauch, F. Tung, and C.~T. Striebel.
\newblock Maximum likelihood estimates of linear dynamic systems.
\newblock \emph{AIAA Journal}, 3\penalty0 (8):\penalty0 1445--1450, 1965.

\bibitem[Ravi et~al.(2024)Ravi, Gabeur, Hu, Hu, Ryali, Ma, Khedr, R{\"a}dle,
  Rolland, Gustafson, Mintun, Pan, Alwala, Carion, Wu, Girshick, Doll{\'a}r,
  and Feichtenhofer]{ravi2024sam2}
Nikhila Ravi, Valentin Gabeur, Yuan-Ting Hu, Ronghang Hu, Chaitanya Ryali,
  Tengyu Ma, Haitham Khedr, Roman R{\"a}dle, Chloe Rolland, Laura Gustafson,
  Eric Mintun, Junting Pan, Kalyan~Vasudev Alwala, Nicolas Carion, Chao-Yuan
  Wu, Ross Girshick, Piotr Doll{\'a}r, and Christoph Feichtenhofer.
\newblock {SAM} 2: {Segment Anything} in images and videos, 2024.
\newblock arXiv:2408.00714.

\bibitem[Ren et~al.(2024)Ren, Liu, Zeng, Lin, Li, Cao, Chen, Huang, Chen, Yan,
  Zeng, Zhang, Li, Yang, Li, Jiang, and Zhang]{ren2024grounded}
Tianhe Ren, Shilong Liu, Ailing Zeng, Jing Lin, Kunchang Li, He Cao, Jiayu
  Chen, Xinyu Huang, Yukang Chen, Feng Yan, Zhaoyang Zeng, Hao Zhang, Feng Li,
  Jie Yang, Hongyang Li, Qing Jiang, and Lei Zhang.
\newblock Grounded {SAM}: Assembling open-world models for diverse visual
  tasks, 2024.
\newblock arXiv:2401.14159.

\bibitem[Skvrna and Neumann(2025)]{skvrna2025monosowa}
Jan Skvrna and Lukas Neumann.
\newblock {MonoSOWA}: Scalable monocular {3D} object detector without human
  annotations, 2025.
\newblock arXiv:2501.09481.

\bibitem[Sun et~al.(2025)Sun, Liu, He, Tian, and Wang]{sun2025annofreeod}
Boyi Sun, Yuhang Liu, Houxin He, Yonglin Tian, and Fei-Yue Wang.
\newblock {AnnofreeOD}: Detecting all classes at low frame rates without human
  annotations.
\newblock In \emph{ICCV}, pages 5315--5325, 2025.

\bibitem[Sun et~al.(2020)Sun, Kretzschmar, Dotiwalla, Chouard, Patnaik, Tsui,
  Guo, Zhou, Chai, Caine, Vasudevan, Han, Ngiam, Zhao, Timofeev, Ettinger,
  Krivokon, Gao, Joshi, Zhao, Cheng, Zhang, Shlens, Chen, and
  Anguelov]{sun2019scalability}
Pei Sun, Henrik Kretzschmar, Xerxes Dotiwalla, Aurelien Chouard, Vijaysai
  Patnaik, Paul Tsui, James Guo, Yin Zhou, Yuning Chai, Benjamin Caine, Vijay
  Vasudevan, Wei Han, Jiquan Ngiam, Hang Zhao, Aleksei Timofeev, Scott
  Ettinger, Maxim Krivokon, Amy Gao, Aditya Joshi, Sheng Zhao, Shuyang Cheng,
  Yu Zhang, Jonathon Shlens, Zhifeng Chen, and Dragomir Anguelov.
\newblock Scalability in perception for autonomous driving: {Waymo} open
  dataset.
\newblock In \emph{CVPR}, pages 2443--2451, 2020.
\newblock arXiv:1912.04838.

\bibitem[Tempfli et~al.(2026)Tempfli, Rivera, and Lienkamp]{tempfli2026vespa}
Levente Tempfli, Esteban Rivera, and Markus Lienkamp.
\newblock {VESPA}: Open-world auto-labeling for {3D} object detection in
  autonomous driving.
\newblock In \emph{CVPR Findings}, 2026.
\newblock arXiv:2507.20397.

\bibitem[Varma et~al.(2019)Varma, Subramanian, Namboodiri, Chandraker, and
  Jawahar]{varma2019idd}
Girish Varma, Anbumani Subramanian, Anoop Namboodiri, Manmohan Chandraker, and
  C.~V. Jawahar.
\newblock {IDD}: A dataset for exploring problems of autonomous navigation in
  unconstrained environments.
\newblock In \emph{WACV}, 2019.

\bibitem[Vasa et~al.(2026)Vasa, Ramadwar, Darabattula, Anwar, Antol, Vatavu,
  Monninger, and Ding]{vasa2026autovdc}
Santosh Vasa, Aditi Ramadwar, Jnana Rama~Krishna Darabattula, Md~Zafar Anwar,
  Stanislaw Antol, Andrei Vatavu, Thomas Monninger, and Sihao Ding.
\newblock {AutoVDC}: Automated vision data cleaning using vision-language
  models.
\newblock In \emph{IV Workshops}, pages 2167--2173, 2026.

\bibitem[Wei et~al.(2021)Wei, Su, Lu, and Zhou]{wei2021fgr}
Yi Wei, Shang Su, Jiwen Lu, and Jie Zhou.
\newblock {FGR}: Frustum-aware geometric reasoning for weakly supervised {3D}
  vehicle detection.
\newblock In \emph{ICRA}, pages 4348--4354, 2021.
\newblock arXiv:2105.07647.

\bibitem[Weng et~al.(2020)Weng, Wang, Held, and Kitani]{weng2020ab3dmot}
Xinshuo Weng, Jianren Wang, David Held, and Kris Kitani.
\newblock {3D} multi-object tracking: A baseline and new evaluation metrics.
\newblock In \emph{IROS}, 2020.
\newblock arXiv:1907.03961.

\bibitem[Wilson et~al.(2021)Wilson, Qi, Agarwal, Lambert, Singh, Khandelwal,
  Pan, Kumar, Hartnett, {Kaesemodel Pontes}, Ramanan, Carr, and
  Hays]{wilson2023argoverse}
Benjamin Wilson, William Qi, Tanmay Agarwal, John Lambert, Jagjeet Singh,
  Siddhesh Khandelwal, Bowen Pan, Ratnesh Kumar, Andrew Hartnett, Jhony
  {Kaesemodel Pontes}, Deva Ramanan, Peter Carr, and James Hays.
\newblock Argoverse 2: Next generation datasets for self-driving perception and
  forecasting.
\newblock In \emph{NeurIPS Datasets and Benchmarks}, 2021.
\newblock arXiv:2301.00493.

\bibitem[Yang et~al.(2023)Yang, Casas, Dvornik, Segal, Xiong, Hu, Fang, and
  Urtasun]{yang2023labelformer}
Anqi~Joyce Yang, Sergio Casas, Nikita Dvornik, Sean Segal, Yuwen Xiong, Jordan
  Sir~Kwang Hu, Carter Fang, and Raquel Urtasun.
\newblock {LabelFormer}: Object trajectory refinement for offboard perception
  from {LiDAR} point clouds.
\newblock In \emph{CoRL}, 2023.
\newblock arXiv:2311.01444.

\bibitem[Yin et~al.(2021)Yin, Zhou, and Kr{\"a}henb{\"u}hl]{yin2021center}
Tianwei Yin, Xingyi Zhou, and Philipp Kr{\"a}henb{\"u}hl.
\newblock Center-based {3D} object detection and tracking.
\newblock In \emph{CVPR}, pages 11784--11793, 2021.
\newblock arXiv:2006.11275.

\bibitem[Yu et~al.(2020)Yu, Chen, Wang, Xian, Chen, Liu, Madhavan, and
  Darrell]{yu2018bdd100k}
Fisher Yu, Haofeng Chen, Xin Wang, Wenqi Xian, Yingying Chen, Fangchen Liu,
  Vashisht Madhavan, and Trevor Darrell.
\newblock {BDD100K}: A diverse driving dataset for heterogeneous multitask
  learning.
\newblock In \emph{CVPR}, pages 2633--2642, 2020.

\bibitem[Zhang et~al.(2017)Zhang, Xu, Dong, and Dolan]{zhang2017efficient}
Xiao Zhang, Wenda Xu, Chiyu Dong, and John~M. Dolan.
\newblock Efficient {L}-shape fitting for vehicle detection using laser
  scanners.
\newblock In \emph{IV}, pages 54--59, 2017.

\bibitem[Zunair et~al.(2024)Zunair, Khan, and Hamza]{zunair2024rsud20k}
Hasib Zunair, Shakib Khan, and A.~Ben Hamza.
\newblock {RSUD20K}: A dataset for road scene understanding in autonomous
  driving, 2024.
\newblock arXiv:2401.07322.

\end{thebibliography}
}
\clearpage
\pdfximage{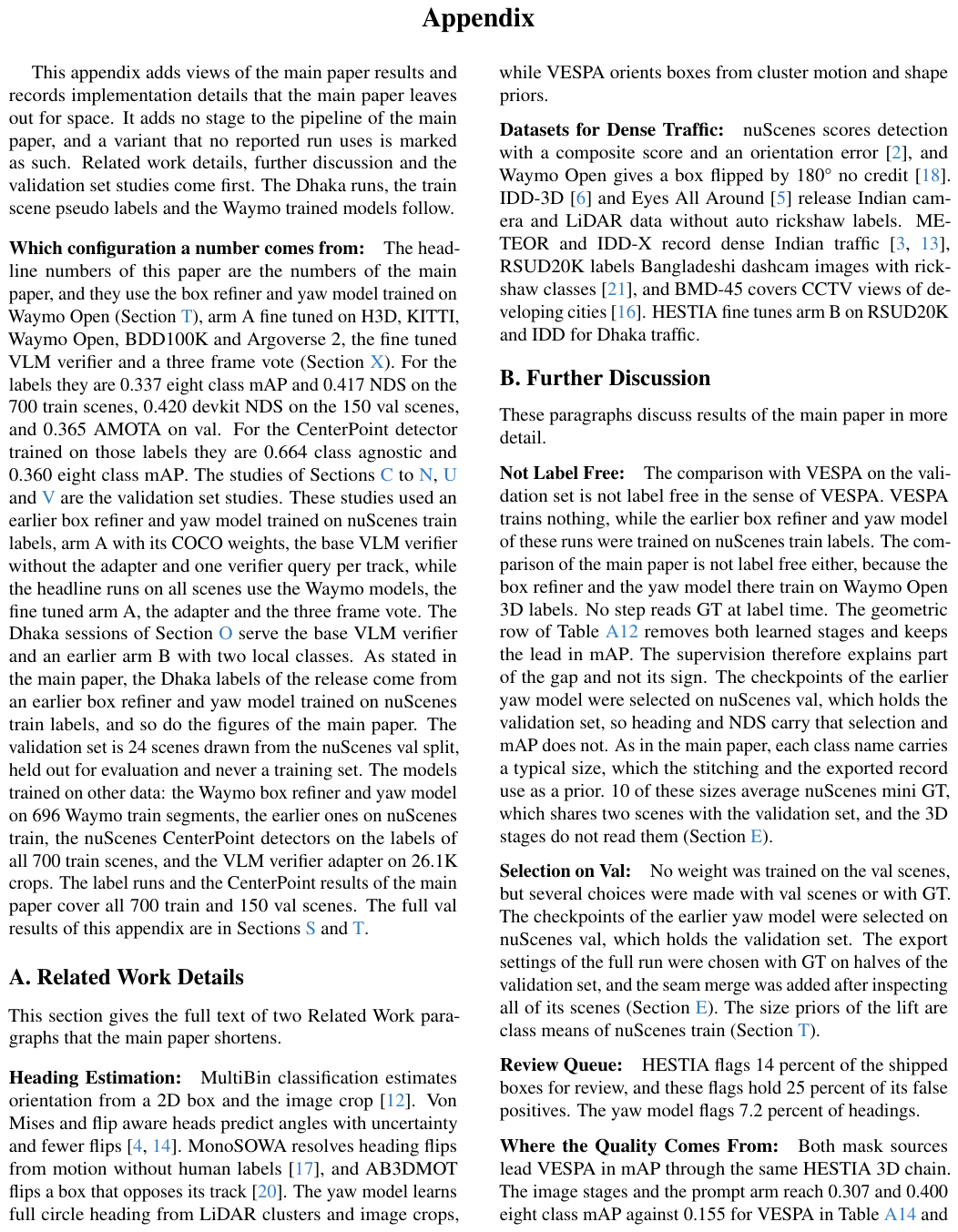}\edef\suppn{\the\pdflastximagepages}
\newcount\supppage \supppage=1
\pagestyle{empty}
\AddToShipoutPicture{\AtPageLowerLeft{\includegraphics[page=\the\supppage,width=\paperwidth,height=\paperheight]{supplementary.pdf}}}
\loop \null\thispagestyle{empty}\clearpage \advance\supppage by 1 \ifnum\supppage>\suppn\relax\else\repeat
\end{document}